\documentclass[10pt,twocolumn,letterpaper]{article}
\newif\ifarxiv%
\arxivtrue
\usepackage{microtype}
\usepackage[pagenumbers]{iccv} %

\DeclareMathOperator*{\argmin}{arg\,min}
\definecolor{HOPS}{RGB}{31, 119, 180}

\usepackage{booktabs} %
\usepackage{xcolor} %
\usepackage{graphicx} %
\usepackage{array} %
\usepackage{fix-cm}

\definecolor{iccvblue}{rgb}{0.21,0.49,0.74}
\usepackage[pagebackref,breaklinks,colorlinks,allcolors=iccvblue]{hyperref}

\def\paperID{12849} %
\def\confName{ICCV}
\def\confYear{2025}

\title{A \textcolor{HOPS}{H}yperdimensional \textcolor{HOPS}{O}ne \textcolor{HOPS}{P}lace \textcolor{HOPS}{S}ignature to Represent Them All:\\Stackable Descriptors For Visual Place Recognition}
\ifarxiv
    \author{Connor Malone \ \ \ 
    Somayeh Hussaini \ \ \ 
    Tobias Fischer \ \ \ 
    Michael Milford\\
    QUT Centre for Robotics, Queensland University of Technology, Australia\\
    {\tt\small \{cj.malone, s.hussaini, tobias.fischer, michael.milford\}\tt\small @qut.edu.au}
    }
\else
    \author{Connor Malone\\
    {\tt\small cj.malone@qut.edu.au}
    \and
    Somayeh Hussaini\\
    {\tt\small s.hussaini@qut.edu.au}
    \and
    Tobias Fischer\\
    {\tt\small tobias.fischer@qut.edu.au}
    \and
    Michael Milford\\
    {\tt\small michael.milford@qut.edu.au}
    }
\fi

\ifarxiv
    \usepackage{fancyhdr}
    \fancypagestyle{smallHeaders}{%
        \fancyhf{}
        \fancyhead[C]{ \fontsize{10}{10}\selectfont Preprint version: Accepted to ICCV 2025. Final version available at (TBC)}
        \fancyfoot[C]{\thepage}
    }%
    
    \fancypagestyle{bigHeaders}{%
        \fancyhf{}
        \fancyhead[C]{ \fontsize{12}{12}\selectfont Preprint version: Accepted to ICCV 2025. Final version available at (TBC)}
        
        \fancyfoot[C]{\thepage}
    }%
\fi

\begin{document}
    \maketitle
    \ifarxiv
        \thispagestyle{bigHeaders}
        \pagestyle{smallHeaders}
    \fi
    \begin{abstract}
    Visual Place Recognition (VPR) enables coarse localization by comparing query images to a reference database of geo-tagged images. Recent breakthroughs in deep learning architectures and training regimes have led to methods with improved robustness to factors like environment appearance change, but with the downside that the required training and/or matching compute scales with the number of distinct environmental conditions encountered. Here, we propose \textcolor{HOPS}{H}yperdimensional \textcolor{HOPS}{O}ne \textcolor{HOPS}{P}lace \textcolor{HOPS}{S}ignatures (\textcolor{HOPS}{HOPS}) to simultaneously improve the performance, compute and scalability of these state-of-the-art approaches by fusing the descriptors from multiple reference sets captured under different conditions. \textcolor{HOPS}{HOPS} scales to any number of environmental conditions by leveraging the Hyperdimensional Computing framework. Extensive evaluations demonstrate that our approach is highly generalizable and consistently improves recall performance across all evaluated VPR methods and datasets by large margins. Arbitrarily fusing reference images without compute penalty enables numerous other useful possibilities, three of which we demonstrate here: descriptor dimensionality reduction with no performance penalty, stacking synthetic images, and coarse localization to an entire traverse or environmental section.

\end{abstract}    
    \section{Introduction}
\label{sec:intro}

    Localization is a critical task in robotics, autonomous vehicles~\cite{chalvatzaras2022survey, kumar2023survey}, and augmented reality~\cite{sarlin2022lamar, plizzari2024outlook}. Long-term operation requires localization systems that are robust to factors like lighting, weather and dynamic scene changes –– all of which significantly impact a place's appearance~\cite{toft2020long}.
    
    Visual Place Recognition (VPR) is the task of identifying previously visited places given a query image and a database of geo-tagged reference images~\cite{lowry2015visual, zhang2021visual, masone2021survey, garg2021your, SchubertRAM2023ICRA2024}. In applications such as loop closure in Simultaneous Localization and Mapping (SLAM)~\cite{durrant2006simultaneous, cadena2016past, tsintotas2022revisiting}, VPR is often formulated as an image retrieval problem that provides coarse localization estimates, which are then refined in a hierarchical process using feature matching approaches~\cite{murillo2007omnidirectional, sarlin2018leveraging, sarlin2019coarse}. %

    \begin{figure}
        \centering
        \includegraphics[width=\linewidth]{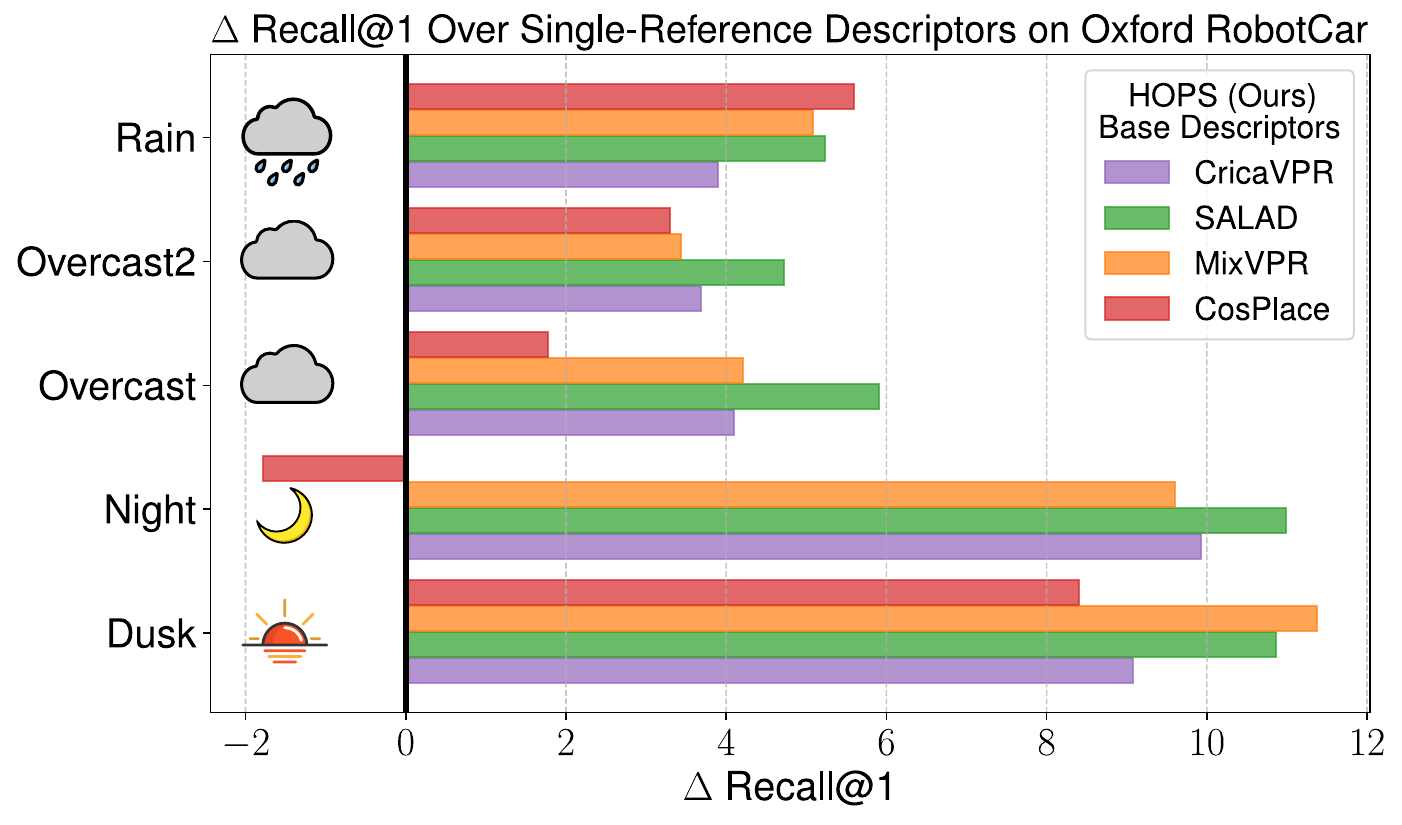}
        \caption{Here, we demonstrate near unanimous improvements to recall@1 by using our \textcolor{HOPS}{HOPS} fused descriptors. We show improvement values relative to the best recall achieved by a single reference set, while our approach uses hyperdimensional computing to fuse descriptors from multiple reference sets without increasing dimensionality. %
        }
        \label{fig:frontPage}
    \end{figure}

    Most state-of-the-art (SOTA) VPR methods use deep learning models to represent images as feature-based descriptors~\cite{ali2023mixvpr, berton2023eigenplaces, keetha2023anyloc, izquierdo2024optimal, lu2024cricavpr}. While significant progress towards VPR that is robust to lighting, weather, viewpoint and other appearance changes has been made, most approaches adopt the general formulation of using a single reference set (often captured in `ideal daytime' conditions) to perform place recognition. To further improve appearance invariance, recent deep learning methods have used multi-condition training sets~\cite{warburg2020mapillary, ali2022gsv}, explicit consideration of multiple instances of places captured under varying conditions to improve feature robustness~\cite{berton2023eigenplaces, lu2024cricavpr}, and domain adaptation~\cite{berton2021adaptive, hu2020dasgil}. Further work has attempted to consolidate separate VPR matches across multiple reference datasets~\cite{fischer2020event, molloy2020intelligent}, or simply develop ever more robust feature extractors~\cite{keetha2023anyloc, berton2023eigenplaces, lu2024cricavpr, lu2024towards, tzachor2024effovpr}.   
    
    In this work, we explore an alternative approach for improving general robustness to appearance changes which does not involve computationally- and time-intensive training of a new deep learned feature extractor (see Figure~\ref{fig:frontPage}). 
    We instead propose \textcolor{HOPS}{H}yperdimensional \textcolor{HOPS}{O}ne \textcolor{HOPS}{P}lace \textcolor{HOPS}{S}ignatures (\textcolor{HOPS}{HOPS})\footnote{https://github.com/CMalone-Jupiter/HOPS} to fuse VPR descriptors from the same place captured under varying conditions using the Hyperdimensional Computing (HDC) framework~\cite{neubert2019introduction, kleyko2023survey} -- as opposed to fusing VPR descriptors obtained by complementary techniques~\cite{neubert2021hyperdimensional}. 
    
    This leverages the capability of current SOTA VPR descriptors to match images in similar domains whilst using the HDC formulation to avoid any additional training, and computational or memory costs. Importantly, \textcolor{HOPS}{HOPS} is generalizable and complementary to existing SOTA VPR descriptors. We make the following contributions:
    \begin{enumerate}
        \item The first use of a Hyperdimensional Computing (HDC) framework for fusing multiple reference sets---either from different traverses of the environment, or synthetically generated using image augmentations---in VPR to improve robustness to appearance changes without increasing computation or memory requirements.
        \item Extensive experiments showing the framework generalizes across several SOTA VPR methods and multiple datasets with various challenging condition changes, generally outperforming the best single reference set by large margins and achieving better performance than other multi-reference set approaches that require additional computation or memory costs.  %

        \item An alternative operation mode that significantly reduces computation and memory costs without performance cost: in the case of high-dimensional descriptors such as SALAD~\cite{izquierdo2024optimal} (8448D) and CricaVPR~\cite{lu2024cricavpr} (10752D) about a $97\%$ and $95\%$ reduction in feature dimensions, respectively, and for low-dimensional descriptors such as CosPlace~\cite{berton2022rethinking} and EigenPlaces~\cite{berton2023eigenplaces} (both 512D) still achieving about a $50\%$ and $25\%$ reduction, respectively.
    \end{enumerate}

    \noindent The ability to combine reference images together for Visual Place Recognition to improve recognition performance without any scaling of matching compute is a powerful and versatile one. Here, our primary demonstration is in showing the performance and compute benefits in appearance-invariant place matching, through both combining multiple real-world reference traverses and synthetic imagery. Such a capability opens up many other possibilities, some of which are discussed in Section~\ref{sec::conclusion}.

    \section{Related Work}
\label{sec:related_work}

    \subsection{Visual Place Recognition}
    \label{subsec::related_vpr}

    In Visual Place Recognition (VPR), images are typically converted to high-level feature descriptors robust to appearance and viewpoint changes, allowing a query image to match the correct reference image in the feature space~\cite{SchubertRAM2023ICRA2024, masone2021survey}. 
    Early VPR solutions used handcrafted feature descriptors, including global aggregation methods such as Bag of Words (BoW)~\cite{sivic2003video, csurka2004visual}, Fischer Vectors (FV)~\cite{perronnin2007fisher, perronnin2010large}, Vector of Locally Aggregated Descriptors (VLAD)~\cite{jegou2010aggregating, arandjelovic2013all}, and local descriptors such as SIFT~\cite{Lowe1999SIFT} and SURF~\cite{bay2008speeded}. With deep learning, these methods evolved into architectures such as NetVLAD~\cite{arandjelovic2016netvlad}, NetBoW~\cite{radenovic2016cnn}, and NetFV~\cite{miech2017learnable}.
    Since the introduction of CNNs to VPR~\cite{arandjelovic2016netvlad}, deep learning techniques have enabled greater robustness against appearance and viewpoint changes, which include works such as DELF~\cite{noh2017large}, DELG~\cite{Cao2020DELG}, DOLG~\cite{yang2021dolg} %
    and \mbox{SuperGlue~\cite{sarlin2020superglue}}. 
    Recent approaches address VPR challenges through  
    spatial pooling and aggregation methods such as Generalized Mean Pooling (GeM)~\cite{radenovic2018fine}, and Conv-AP~\cite{ali2022gsv},  
    innovative architectures~\cite{ali2023mixvpr}, 
    VPR-method-agnostic feature alignment procedures such as MeshVPR~\cite{berton2024meshvpr},
    effective training regimes~\cite{trivigno2023divide, berton2022rethinking, berton2023eigenplaces}, and targeted VPR-specific loss functions~\cite{revaud2019learning,leyva2023data}. 
    MixVPR~\cite{ali2023mixvpr} uses CNN backbones and Feature Mixer layers to establish global relationships within feature maps. 
    EigenPlaces~\cite{berton2023eigenplaces} targets viewpoint tolerance by dividing the training dataset to form small classes with images of multiple perspectives. 
    CosPlace~\cite{berton2022rethinking} reformulates VPR training as a classification task by organizing data into geographically distinct classes. 
    Generalized Contrastive Loss (GCL)~\cite{leyva2023data} improves global descriptor robustness by computing graded similarity for image pairs.

    Other SOTA VPR models leverage vision transformers~\cite{dosovitskiy2021an, han2022survey} for enhanced feature extraction, including \mbox{DinoV2} SALAD~\cite{izquierdo2024optimal} that treats descriptor aggregation as an optimal transport problem, AnyLoc~\cite{keetha2023anyloc} that also uses DinoV2 without VPR-specific fine-tuning, CricaVPR~\cite{lu2024cricavpr} that introduces cross-image correlation awareness, and BoQ~\cite{ali2024boq} which learns a set of global queries, using cross-attention with local input features to derive global representations.

    Other VPR approaches enhance performance by developing two-stage retrieval techniques, which initially identify top-$k$ candidates using global features, and then re-ranks these candidates using local features~\cite{masone2021survey}. Recent two-stage approaches include Patch-NetVLAD~\cite{hausler2021patch} and transformer-based methods such as TransVPR~\cite{wang2022transvpr}, ETR~\cite{zhang2023etr}, 
    $R^{2}$Former~\cite{zhu2023r2former}, SelaVPR~\cite{lu2024towards}, and \mbox{EffoVPR}~\cite{tzachor2024effovpr}. Relevant to this work, \cite{barbarani2023local} investigates how existing local features and re-ranking methods can be used to improve VPR with challenges such as night time conditions and image occlusions.

    \subsection{Multi-Reference and Fusion Approaches}
    \label{subsec::related_fusion}

    Several VPR techniques focus on fusion approaches~\cite{jacobson2015autonomous, siva2018omnidirectional, hou2023fe, xin2019real, yu2019spatial} 
    or consider multiple reference sets~\cite{churchill2012practice, linegar2015work, molloy2020intelligent, vysotska2019effective} 
    by generating enriched reference maps that enable robots to perform long-term autonomous navigation as changes in the environment over time can be incorporated~\cite{SchubertRAM2023ICRA2024}. 
    Feature fusion has been used to
    fuse input data from a range of sensors such as camera, laser and sonar~\cite{jacobson2015autonomous},
    omnidirectional observations with a depth sensor and camera~\cite{siva2018omnidirectional}, and image-based and event-based camera data~\cite{hou2023fe}. 
    Feature fusion has also been used for re-ranking top-candidate matches obtained through matching global feature descriptors~\cite{xin2019real, yu2019spatial}. 

    Training using multi-condition datasets is a common way for VPR methods to achieve more invariant features~\cite{warburg2020mapillary, ali2022gsv}. While not strictly using multiple reference sets, the SOTA VPR method CricaVPR even specifically incorporates correlations between images of the same place captured under varying conditions~\cite{lu2024cricavpr}.
    
    Multiple reference sets have been more explicitly used for improving place recognition performance by incrementally adapting to appearance changes~\cite{churchill2012practice} and using probabilistic approaches to predict the best reference set to use for a given query image~\cite{linegar2015work, molloy2020intelligent}.
    \cite{vysotska2019effective} used an efficient hashing technique to generate feature descriptors and used a data association graph to store representations from multiple reference sets, and performed place matching using an informed search.
    While these works~\cite{churchill2012practice, linegar2015work, molloy2020intelligent, vysotska2019effective} have addressed the problem of multiple reference maps, an on-going concern is the increasing storage and computational requirements with increase in the number of reference sets.

    \subsection{Hyperdimensional Computing Frameworks}
    \label{subsec:related_HDC}

    Hyperdimensional Computing (HDC), also known as Vector Symbolic Architectures (VSA), is a brain-inspired computing framework~\cite{ge2020classification, karunaratne2020memory}. 
    HDC is used to handle data which is represented in extremely high, or `hyper', dimensional spaces~\cite{ge2020classification}; expected to have thousands or tens of thousands dimensions. One of the key properties in such hyperdimensional spaces is that there is a high likelihood that two randomly sampled vectors will be near or `quasi' orthogonal to one another~\cite{schlegel2022comparison}. As a result, several HDC operations can be performed to improve the computational and memory efficiency of dealing with these vectors, including bundling, binding, and permutation~\cite{ge2020classification}.
    
    Of interest for this paper is bundling, which fuses sets of input vectors such that the output vector is similar to all input vectors~\cite{neubert2021hyperdimensional}. One method for bundling which has precedence in VPR literature is an element-wise sum of the vectors~\cite{neubert2021hyperdimensional}.
    The binding operation can be used to assign `role' or `class' information to vectors. The output of binding is not similar to the two input vectors but can be reversed to recover the input components; one implementation  is through an element-wise multiplication of two vectors~\cite{neubert2021hyperdimensional}. 

    HDC has been used in a range of machine learning applications for learning temporal patterns such as text classification~\cite{kleyko2018classification}, addressing catastrophic forgetting in deep learning-based architectures~\cite{cheung2019superposition}, in robotics for reactive behavior learning, and object and place recognition~\cite{neubert2019introduction}, and out-of-distribution detection~\cite{wilson2023hyperdimensional}. 

    In the context of VPR, \cite{neubert2021vector} presented the Vector Semantic Representations (VSR) image descriptor, which uses HDC to encode the appearance and semantic properties of a place, as well as the topological relationship between semantic classes.  
    \cite{neubert2021hyperdimensional} presented an HDC-based framework to aggregate image descriptors from multiple different global VPR methods, or for aggregating local features and binding their image position information. \cite{neubert2021hyperdimensional} exploits the HDC properties of orthogonal vectors to fuse descriptors from different VPR methods -- we differ from this by instead exploiting the reinforcement of features by fusing multiple reference descriptors of the same place from the same VPR method.
    \section{Methodology}
\label{sec::method}

\subsection{Visual Place Recognition Formulation}
We formulate Visual Place Recognition (VPR) as an image retrieval task. Given a query image of the current place and a database of geo-tagged reference images, our goal is to identify the reference image that most closely resembles the query. State-of-the-art VPR methods commonly use deep neural networks to embed images as $n$-dimensional feature vectors, thereby abstracting complex visual scenes into compact representations.

Formally, let $\mathbf{q} \in \mathbb{R}^n$ represent the feature vector of the query image and $\mathbf{R} = \{\mathbf{r}_i\}$ the set of geo-tagged reference vectors, with $\mathbf{r}_i \in \mathbb{R}^n$ and $|\mathbf{R}|=M$ being the number of reference images. To compute the degree of similarity between the query and each reference, we calculate a distance vector $\mathbf{d} = [d(\mathbf{q}, \mathbf{r}_1), d(\mathbf{q}, \mathbf{r}_2), \dots, d(\mathbf{q}, \mathbf{r}_M)]$, where $d(\cdot)$ denotes the cosine distance. The estimated location is then derived by selecting the reference with the minimum distance:

\begin{equation}
    \mathbf{r}_{\text{match}} = \argmin_i{d(\mathbf{q}, \mathbf{r}_i)}.
\end{equation}

This approach critically depends on the robustness of neural network feature extractors, which must maintain discriminative power across various environmental conditions and viewpoints for each unique place. Achieving high consistency across such changes is crucial for robust and long-term VPR. However, instead of relying solely on improved feature extraction, we propose leveraging Hyperdimensional Computing (HDC) to fuse multiple reference sets into \textcolor{HOPS}{H}yperdimensional \textcolor{HOPS}{O}ne \textcolor{HOPS}{P}lace \textcolor{HOPS}{S}ignatures (\textcolor{HOPS}{HOPS}), enhancing condition invariance without altering existing VPR descriptors.

\subsection{Bundling Reference Datasets}
\label{subsec::bundling}
Our approach exploits the properties of high-dimensional spaces by aggregating multiple feature vectors to create a fused descriptor which is similar to all inputs. In other words, we put forward the idea that hyperdimensional feature vectors from the same place, captured under different conditions, can be combined to form a unified descriptor that remains robust against minor variations.

Formally, let $\mathbf{r}^k$ be feature vectors representing the same place under different conditions $k$, with an additional noise vector $\mathbf{z}$ affecting either vector. Due to quasi-orthogonality, the influence of $\mathbf{z}$ on the cosine similarity between $\mathbf{r}^l$ and $\mathbf{r}^m$ ($l \neq m$) is negligible in high-dimensional space, preserving the similarity despite the noise.
$\mathbf{r}_{\text{fused},i}$ combines $K$ reference descriptors from the same place $i$ across diverse conditions, allowing salient features to reinforce while diminishing transient ones:
\vspace{-\baselineskip}
\begin{equation}
    \mathbf{r}_{\text{fused},i} = \sum_{k=1}^K \mathbf{r}_i^k.
\end{equation}
\vspace{-\baselineskip}

Bundling via summing has the useful property of being able to `stack' many reference descriptors, which is useful as new descriptors can be easily added to the fusion over time as the places are revisited. It maintains a complexity of $\mathcal{O}(M)$.

\subsection{Gaussian Random Projection}
\label{subsec::GRP}

 Beyond the core benefits of our \textcolor{HOPS}{HOPS} approach for fusing descriptors without additional compute or memory overhead, it also enables other beneficial applications such as improved performance after dimensionality reduction operations. To demonstrate this, we use Gaussian Random Projection as a representative method in an additional experiment (Section~\ref{subsec:red-dim_results}) to project feature vectors into a lower-dimensional space. Using a random projection matrix, the Johnson-Lindenstrauss Lemma asserts that the distance between a set of points in high-dimensional space can be approximately preserved when embedding in a lower-dimensional space~\cite{achlioptas2003database}. In this work, we use Gaussian Random Projections to evaluate the capacity for \textcolor{HOPS}{HOPS} to reduce the descriptor dimensionality required to maintain performance. This is \textbf{not} done for core experimental results (Tables~\ref{table:HDC_refs_RobotCar}-\ref{table:RobotCar_Synth} and Figures~\ref{fig:recall_prog}-\ref{fig:error_hist}).

Given a high-dimensional feature vector $\mathbf{r}_{\text{fused},i} \in \mathbb{R}^n$, the Gaussian Random Projection $\mathbf{G} \in \mathbb{R}^{o \times n}$ projects $\mathbf{r}_{\text{fused},i}$ to a lower-dimensional space $\mathbb{R}^o$ where $o \ll n$. The projection is performed using matrix multiplication:

\vspace{-\baselineskip}
\begin{equation}
\mathbf{{\hat{r}}}_{\text{fused},i} = \mathbf{G} \mathbf{r}_{\text{fused},i},
\end{equation}
\vspace{-\baselineskip}

where elements in $\mathbf{G}$ are sampled from a Gaussian distribution $\mathcal{N}(0, \frac{1}{n})$, and $\mathbf{{\hat{r}}}_{\text{fused},i} \in \mathbb{R}^o$ is the lower-dimensional representation of $\mathbf{r}_{\text{fused},i}$.

    \section{Experiments}
    \label{sec::experiments}
    This section first details the experimental setup (Section~\ref{subsec:exp_setup}), including the datasets, underlying VPR descriptors, and metrics used to evaluate \textcolor{HOPS}{HOPS}. Section~\ref{subsec:multirefbaseline} introduces two strong baseline multi-reference approaches. We then provide experimental results and analysis for place matching performance, including comparison to single-set baselines (Section~\ref{subsec:single-ref_results}), multi reference-set baselines (Section~\ref{subsec:multi-ref_results}), and experiments with reduced dimensionality descriptors (Section~\ref{subsec:red-dim_results}). The section ends with a study where image augmentations are used to generate multiple reference sets (Section~\ref{subsec:synth_results}) and another study on dataset identification (Section~\ref{subsec:other_apps}). 

    \subsection{Experimental Setup}
        \label{subsec:exp_setup}
        \textbf{General Setup:} Throughout our experiments, we evaluate VPR performance using a single-stage image retrieval pipeline. That is, for every query descriptor, we create a ranked list from the set of reference descriptors in order from most to least similar.%
        
        \textbf{Datasets:}  
        To demonstrate the applicability and robustness of our approach across diverse real-world environments and conditions, we evaluate results across three datasets~\cite{RobotCarDatasetIJRR, Niko13, Jake2015SFU}, each of which contain images from a unique route captured under varying conditions.  
        The overarching properties of these datasets include urban, suburban, and rural environments captured under various times of day, seasons, weather conditions, and dynamic elements such as structural changes, occlusions, and glare. We also evaluate on the more unstructured \hyperlink{https://www.kaggle.com/datasets/confirm/google-landmark-dataset-v2-micro}{Google Landmarks v2 micro} dataset in the supplemental material.  

        \textit{1) Oxford RobotCar}~\cite{RobotCarDatasetIJRR}: The Oxford RobotCar Dataset contains images from 100 traverses across a route around Oxford  throughout the course of a year, capturing the same places under different lighting conditions due to time of day, in changing weather conditions, and with other dynamic changes. We use six separate traverses: sunny, dusk, night, rainy, and two sets of overcast conditions, following prior works~\cite{molloy2020intelligent, hussaini2022spiking}. Each set contains 3876 images which have been sampled at $\approx$1m intervals and have a direct correlation between sets.

        \textit{2) Nordland}~\cite{Niko13}: The Nordland dataset is often used as a benchmark in VPR literature because it captures a large geographical area of 729km across the four seasons, including a snowy winter and seasonal changes to trees and plants. In this work, we subsample the original image sets to use 3975 images per season, all with direct correlation across sets. As typical in the literature~\cite{hausler2021patch}, we remove stationary periods and tunnel sequences.

        \textit{3) SFU Mountain}~\cite{Jake2015SFU}: The SFU Mountain Dataset provides over 8 hours of sensor data collected with a ClearPath Husky robot on trails around Burnaby Mountain, Canada. We use the following image sets: Dry, Dusk, January, Night, November, September, and Wet. We combine `Part-A' and `Part-B' to provide a single set with 385 images per condition.
        
        \textbf{Baseline VPR Descriptors:} To validate the generalizability and applicability of our approach to SOTA VPR descriptors, we evaluate using a large selection of recent methods: CosPlace~\cite{berton2022rethinking}, EigenPlaces~\cite{berton2023eigenplaces}, MixVPR~\cite{ali2023mixvpr}, \mbox{DinoV2} SALAD~\cite{izquierdo2024optimal} (referred to as SALAD from here on), CricaVPR~\cite{lu2024cricavpr}, and include AnyLoc~\cite{keetha2023anyloc} in the supplemental. For MixVPR~\cite{ali2023mixvpr} and SALAD~\cite{izquierdo2024optimal}, we use the author provided implementations, and for other VPR descriptors, we use the VPR method evaluation repository released with EigenPlaces\footnote{https://github.com/gmberton/VPR-methods-evaluation} which collates the original implementations. We also include NetVLAD~\cite{arandjelovic2016netvlad}, as implemented in the Patch-NetVLAD~\cite{hausler2021patch} repository, as a common benchmark still used in the literature. We re-iterate that techniques such as CricaVPR~\cite{lu2024cricavpr} are trained so that they explicitly consider the correlations between features of the same place under multiple conditions.
        
        \textbf{Evaluation Metrics:} Recall@$N$ is a common and versatile metric commonly used for benchmarking VPR methods. It reports the success rate of a VPR method for retrieving the correct reference image in its top $N$ highest ranked references with respect to similarity with the query.  %
        $N=1$ is mathematically equivalent to the precision at $100\%$ recall~\cite{SchubertRAM2023ICRA2024}. 
        Given the difference in sampling between datasets, we assign the following tolerances, as done in prior works~\cite{xu2020probabilistic, pivovnka2024model, schubert2023visual, hussaini2023applications}, for what are considered true matches: RobotCar, $\pm$ 2 images (which is equivalent to 2m); SFU-Mountain, $\pm$ 1 image; Nordland, $\pm$ 0 images (given the distance between images after subsampling).

    \subsection{Baseline Multi-Reference Approaches}
    \label{subsec:multirefbaseline}
    This section introduces two baseline approaches which have explicit access to multiple reference sets at inference time.

    \textbf{Reference Set Pooling:} A straightforward approach to leveraging multiple reference sets involves pooling all reference images into a single, larger reference set. Given $K$ individual reference sets $\mathbf{r}^k$, this method constructs a unified set $\mathbf{r}_{\text{pooled}} = \bigcup_{k=1}^K \mathbf{r}^k$. During query-time matching, the distance vector $\mathbf{d}_{\text{pooled}}$ is computed by comparing the query vector $\mathbf{q}$ against each feature vector in $\mathbf{r}_{\text{pooled}}$:

    \vspace{-\baselineskip}
    \begin{equation}
     \mathbf{d}_{\text{pooled}} = \left[d(\mathbf{q}, \mathbf{r}^1), d(\mathbf{q}, \mathbf{r}^2), \dots, d(\mathbf{q}, \mathbf{r}^{M\cdot K}) \right].
    \end{equation}
    \vspace{-\baselineskip}
    
    This pooling strategy linearly increases the computational complexity with the number of reference sets $K$, resulting in an overall complexity of $\mathcal{O}(K \cdot M)$, where $M$ represents the number of images in each reference set. This increase can significantly impact memory usage and processing time, especially in large-scale environments. However, it maintains simplicity in its setup.
    
    \textbf{Distance Matrix Averaging:} Another multi-reference baseline approach entails performing VPR separately on each reference set and then averaging the resultant distance matrices~\cite{fischer2020event}. For each reference set $\mathbf{r}^k$, an independent distance vector $\mathbf{d}^k$ is computed between the query $\mathbf{q}$ and the reference vectors in $\mathbf{r}^k$:

    \vspace{-\baselineskip}
    \begin{equation}        
    \mathbf{d}^k = \left[d(\mathbf{q}, \mathbf{r}_1^k), d(\mathbf{q}, \mathbf{r}_2^k), \dots, d(\mathbf{q}, \mathbf{r}_M^k)\right].
    \end{equation}
    
    Once each distance vector $\mathbf{d}^k$ has been computed, they are combined by averaging across corresponding distances, producing a final aggregated distance vector $\mathbf{d}_{\text{avg}}$:

    \begin{equation}        
    \mathbf{d}_{\text{avg}} = \frac{1}{K} \sum_{k=1}^K \mathbf{d}^k.
    \end{equation}

    This averaging approach also scales linearly in computational complexity, $\mathcal{O}(K \cdot M)$, as each reference set requires separate matching computations. However, it offers potential for parallelisation, as the VPR matching for each reference set can be executed independently, enabling efficient processing on multi-core or distributed computing systems.~\cite{fischer2020event} also introduced other approaches which we compare to in the supplemental material, however, distance matrix averaging was reported as the highest performing.
    
    \textbf{Summary:} In both approaches, the increased computation and memory requirements limit scalability, particularly in applications requiring real-time performance. Nonetheless, these baseline methods serve as useful comparative approaches, providing insight into the trade-offs associated with managing multiple reference sets in VPR tasks.

    \begin{figure}
        \centering
        \includegraphics[width=\linewidth]{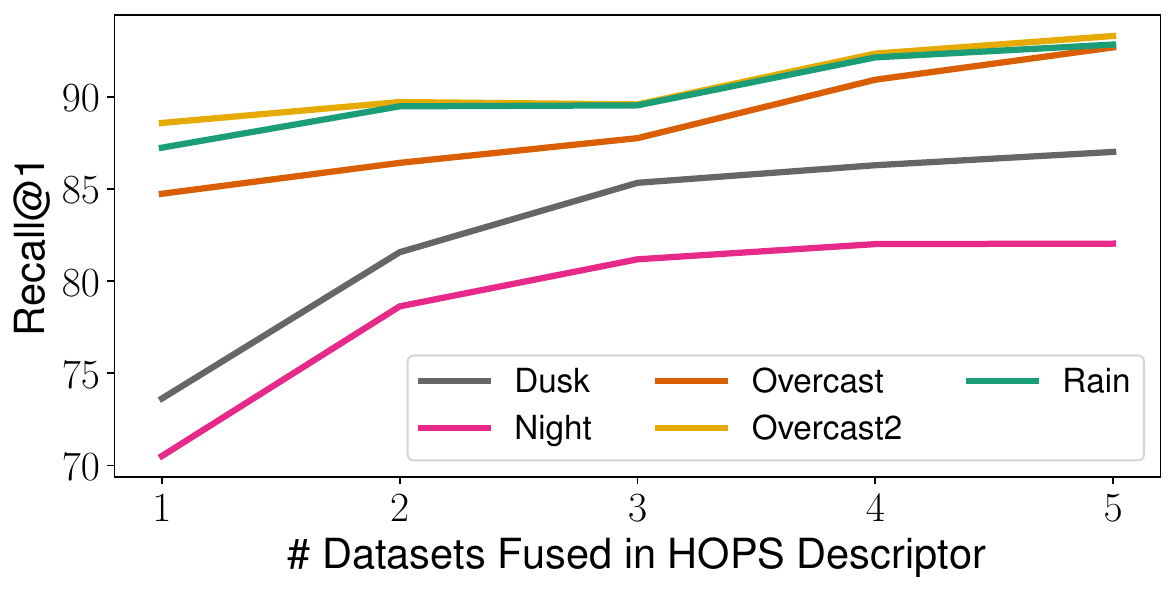}
        \caption{The above plot shows the increase in recall@1 for each Oxford RobotCar query set using our \textcolor{HOPS}{HOPS} descriptors with SALAD as more reference sets are progressively fused. The final fused reference descriptors include all non-query sets.}
        \label{fig:recall_prog}
        \vspace{-2.5\baselineskip}
    \end{figure}

    \subsection{\textbf{\fontsize{10.22pt}{16pt}\selectfont Performance Comparisons to Single Set Baselines}}
    \label{subsec:single-ref_results}

    First, Tables~\ref{table:HDC_refs_RobotCar}-\ref{table:HDC_refs_SFU} demonstrate that using our HDC fused descriptors provide significant performance improvements over the best single reference set baselines. For example, on the Oxford RobotCar dataset, our \textcolor{HOPS}{HOPS} descriptors provide significant improvements to recall, in many cases over absolute $10\%$, even for SOTA VPR descriptors such as SALAD and CricaVPR (Table \ref{table:HDC_refs_RobotCar}). Even on the SFU dataset, where the single reference descriptors already perform strongly, our \textcolor{HOPS}{HOPS} fused descriptors generally improve performance, for example  from $99.0\%$ to $100\%$ on the Dusk query set for SALAD. For the Nordland dataset, \textcolor{HOPS}{HOPS} fused descriptors increased R@1 by absolute 2.9\% on average across the 4 query sets for SALAD. For all experiments, we emphasize that the multi-reference set approaches only combine the sets which are not the query being used for evaluation. Figure~\ref{fig:recall_prog} provides insight into how R@1 is improved incrementally with each additional dataset fused in the \textcolor{HOPS}{HOPS} descriptors, showing the maximum performance occurs with the fusion of all reference sets in this case.
    \setlength\tabcolsep{0.5mm} %

\begin{table*}[!ht]
    \caption{Recall@1 on RobotCar datasets: The table is divided into single-reference and multi-reference approaches. The best single-reference result is \underline{underlined} and the best multi-reference result is \textbf{bolded}. Comparisons in this table should be made vertically down columns. Importantly, our \textcolor{HOPS}{HOPS} fused descriptors are near unanimously better than the best single-reference results (in \textbf{28/30} cases) \textbf{and} better than alternative multi-reference approaches the majority of the time (in \textbf{22/30} cases).}
    \centering
    \scriptsize
    \begin{tabular}{l*{5}{c}|*{5}{c}|*{5}{c}|*{5}{c}|*{5}{c}|*{5}{c}}
        \toprule
        \textbf{Queries $\rightarrow$} 
        & \rotatebox{90}{\raisebox{0.075cm}{\textbf{Dusk}}} & \rotatebox{90}{\raisebox{0.075cm}{\textbf{Night}}} & \rotatebox{90}{\raisebox{0.075cm}{\textbf{Overcast}}} & \rotatebox{90}{\raisebox{0.075cm}{\textbf{Overcast2}}} & \rotatebox{90}{\raisebox{0.075cm}{\textbf{Rain}}} 
        & \rotatebox{90}{\raisebox{0.075cm}{\textbf{Dusk}}} & \rotatebox{90}{\raisebox{0.075cm}{\textbf{Night}}} & \rotatebox{90}{\raisebox{0.075cm}{\textbf{Overcast}}} & \rotatebox{90}{\raisebox{0.075cm}{\textbf{Overcast2}}} & \rotatebox{90}{\raisebox{0.075cm}{\textbf{Rain}}} 
        & \rotatebox{90}{\raisebox{0.075cm}{\textbf{Dusk}}} & \rotatebox{90}{\raisebox{0.075cm}{\textbf{Night}}} & \rotatebox{90}{\raisebox{0.075cm}{\textbf{Overcast}}} & \rotatebox{90}{\raisebox{0.075cm}{\textbf{Overcast2}}} & \rotatebox{90}{\raisebox{0.075cm}{\textbf{Rain}}}
        & \rotatebox{90}{\raisebox{0.075cm}{\textbf{Dusk}}} & \rotatebox{90}{\raisebox{0.075cm}{\textbf{Night}}} & \rotatebox{90}{\raisebox{0.075cm}{\textbf{Overcast}}} & \rotatebox{90}{\raisebox{0.075cm}{\textbf{Overcast2}}} & \rotatebox{90}{\raisebox{0.075cm}{\textbf{Rain}}}
        & \rotatebox{90}{\raisebox{0.075cm}{\textbf{Dusk}}} & \rotatebox{90}{\raisebox{0.075cm}{\textbf{Night}}} & \rotatebox{90}{\raisebox{0.075cm}{\textbf{Overcast}}} & \rotatebox{90}{\raisebox{0.075cm}{\textbf{Overcast2}}} & \rotatebox{90}{\raisebox{0.075cm}{\textbf{Rain}}}
        & \rotatebox{90}{\raisebox{0.075cm}{\textbf{Dusk}}} & \rotatebox{90}{\raisebox{0.075cm}{\textbf{Night}}} & \rotatebox{90}{\raisebox{0.075cm}{\textbf{Overcast}}} & \rotatebox{90}{\raisebox{0.075cm}{\textbf{Overcast2}}} & \rotatebox{90}{\raisebox{0.075cm}{\textbf{Rain}}}\\
        \hline
        \textbf{References} & \multicolumn{5}{c|}{\textbf{NetVLAD (4096D)}} & \multicolumn{5}{c|}{\textbf{SALAD (8448D)}} & \multicolumn{5}{c|}{\textbf{MixVPR (4096D)}} & \multicolumn{5}{c|}{\textbf{CosPlace (512D)}} & \multicolumn{5}{c|}{\textbf{EigenPlaces (512D)}} & \multicolumn{5}{c}{\textbf{CricaVPR (10752D)}}\\
        \hline
        Sunny     & 25.5  & 9.8   & 68.0  & 79.1  & \underline{73.5} & 73.6  & 70.5  & 84.8  & \underline{88.6} & 87.3 & 69.0  & 50.9  & 86.3  & \underline{91.2} & 88.7 &
        44.1 & 14.0 & 78.3 & \underline{86.5} & \underline{84.6} & 42.3 & 13.0 & 81.8 & \underline{88.3} & \underline{87.5} & 81.4 & 77.9 & 90.6 & \underline{93.9} & 92.4\\
        Dusk      & -     & \underline{19.9} & 24.1  & 23.0  & 23.3  & -     & \underline{71.1} & 68.1  & 68.8  & 70.5  & -     & \underline{59.2} & 60.1  & 61.4  & 63.6 &
        - & \underline{21.2} & 42.7 & 42.2 & 44.1 & - & \underline{22.7} & 42.0 & 41.9 & 43.5 & - & 77.8 & 77.2 & 79.4 & 80.8\\
        Night     & 27.6  & -     & 13.6  & 11.6  & 10.6  & 71.7  & -     & 66.4  & 63.7  & 66.2  & 64.6  & -     & 52.2  & 50.3  & 48.3 &
        46.5 & - & 28.8 & 27.2 & 26.5 & 46.3 & - & 25.9 & 25.7 & 22.9 & 81.1 & - & 75.5 & 73.6 & 72.7\\
        Overcast  & \underline{33.0}  & 13.4  & -     & \underline{79.6}  & 72.8  & 74.3  & 71.0  & -     & 88.3  & \underline{87.6} & \underline{71.7}  & 57.2  & -     & 90.6  & \underline{89.6} &
        \underline{48.6} & 18.2 & - & 85.3 & 83.6 & \underline{48.2} & 19.1 & - & 87.9 & 86.9 & \underline{85.7} & \underline{81.0} & - & \underline{93.9} & \underline{93.5}\\
        Overcast2 & 27.0  & 11.2  & \underline{75.9} & -     & 73.2  & 74.4  & 69.1  & \underline{86.8} & -     & 87.2  & 67.4  & 52.0  & \underline{89.1} & -     & 89.5 &
        45.1 & 15.6 & \underline{84.2} & - & 84.2 & 42.7 & 14.0 & \underline{86.5} & - & 86.1 & 84.2 & 77.2 & 92.2 & - & 93.1\\
        Rain      & 29.2  & 9.1   & 68.8  & 72.7  & -     & \underline{76.3}  & 68.6  & 85.8  & 86.6  & -     & 68.3  & 46.0  & 87.1  & 88.8  & - &
        44.8 & 15.0 & 81.6 & 83.9 & - & 44.3 & 15.2 & 84.8 & 86.2 & - & 85.0 & 75.9 & \underline{92.5} & 92.9 & -\\
        \hline
        dMat Avg~\cite{fischer2020event} & 49.0  & 24.6  & 79.4  & 85.9  & 82.6  & 86.2  & 81.3  & 90.2  & 91.4  & 91.5  & 82.9  & \textbf{70.0} & 91.5  & 93.6  & 92.7 &
        56.0 & \textbf{22.9} & 79.8 & 84.6 & 83.7 & \textbf{55.7} & \textbf{23.8} & 84.7 & 88.2 & 86.4 & 94.0 & 89.2 & 95.6 & 96.9 & 96.5\\
        Pooling  & 36.9  & 20.3  & 80.2  & 85.8  & 80.0  & 79.9  & 74.8  & 90.0  & 91.8  & 91.2  & 77.1  & 60.1  & 92.0  & 93.7  & 93.4 &
        55.9 & 21.3 & \textbf{87.8} & \textbf{90.5} & 89.9 & 54.3 & 22.7 & \textbf{90.0} & \textbf{92.1} & \textbf{91.1} & 89.6 & 81.6 & 95.4 & 96.0 & 95.9\\
        \textcolor{HOPS}{HOPS} (Ours) & \textbf{49.8} & \textbf{27.7} & \textbf{83.7} & \textbf{89.5} & \textbf{85.7} & \textbf{87.1} & \textbf{82.1} & \textbf{92.8} & \textbf{93.3} & \textbf{92.9} & \textbf{83.1} & 68.8 & \textbf{93.3} & \textbf{94.7} & \textbf{94.7} &
        \textbf{57.0} & 19.4 & 85.9 & 89.8 & \textbf{90.2} & 54.9 & 20.3 & 89.2 & 92.0 &\textbf{91.1} & \textbf{94.8} & \textbf{91.0} & \textbf{96.6} & \textbf{97.5} & \textbf{97.4}\\
        \bottomrule
    \end{tabular}
    \label{table:HDC_refs_RobotCar}
\end{table*}

    There are three outlier cases where \textcolor{HOPS}{HOPS} fused descriptors perform slightly worse than the best single reference set: using CosPlace or EigenPlaces on Oxford RobotCar Night query (1.8\% and 2.4\% reduction in R@1), and CosPlace on the Nordland Summer query (1.0\% reduction in R@1). As EigenPlaces and CosPlace have relatively low-dimensional descriptors (512D), and HDC principles generally assume vectors have thousands or tens of thousands of dimensions, it is not particularly surprising that the benefits do not hold in these cases. Further investigation may provide insights into how HDC can be applied in these cases.

    \begin{figure}[t]
        \centering
        \setlength\tabcolsep{1.0mm}
        \begin{tabular}{cc}
            \multicolumn{2}{c}{\textbf{SALAD}} \\
            \small{Query: Dusk} & \small{Query: Night} \\
            \includegraphics[width=0.54\linewidth]{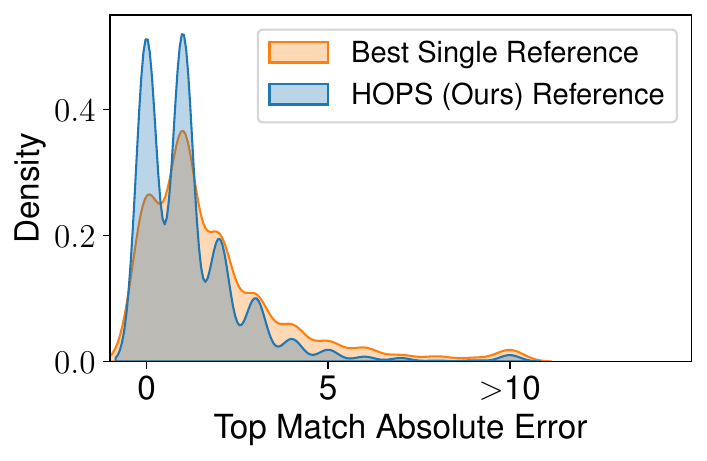} &  
            \includegraphics[width=0.46\linewidth]{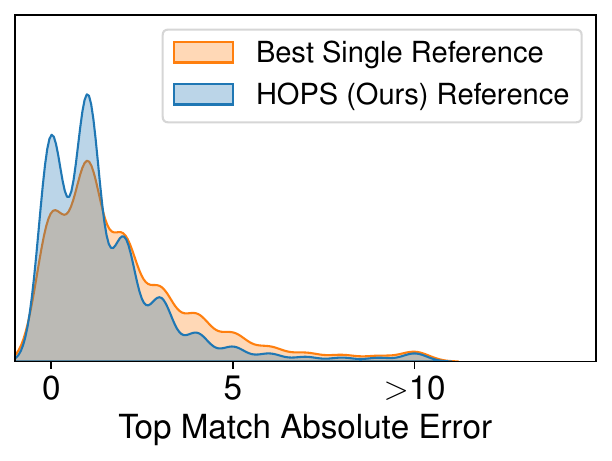} \\
            \small{Query: Overcast} & \small{Query: Rain} \\
            \includegraphics[width=0.54\linewidth]{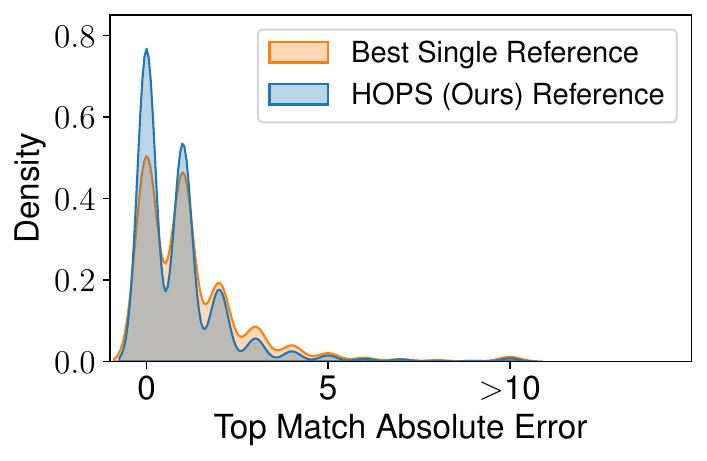} & 
            \includegraphics[width=0.46\linewidth]{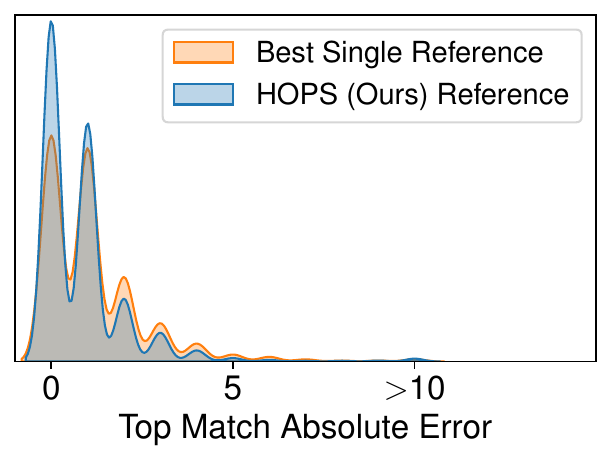} \\
            \multicolumn{2}{c}{\textbf{NetVLAD}} \\
            \small{Query: Dusk} & \small{Query: Rain} \\
            \includegraphics[width=0.54\linewidth]{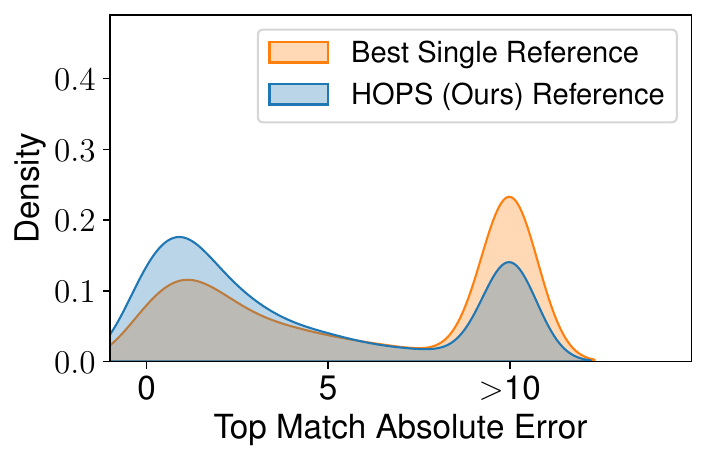} &  
            \includegraphics[width=0.46\linewidth]{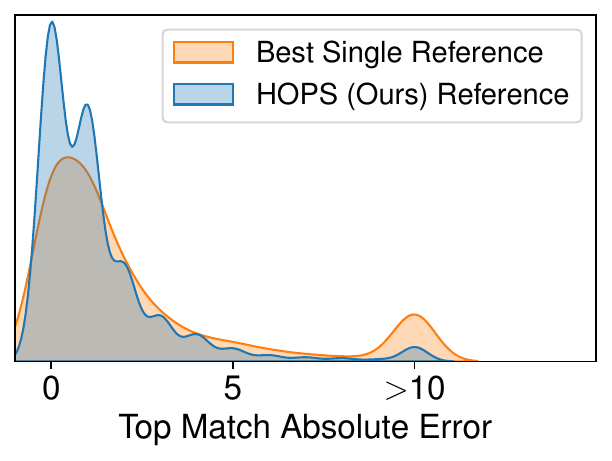} \\
        \end{tabular}
        \caption{\textbf{Top:} Match error density plots for the top VPR match on Oxford RobotCar sets using SALAD descriptors (error measured in frames, $\approx$ 1m/frame for RobotCar). For already high-performing VPR descriptors, our \textcolor{HOPS}{HOPS} fused descriptors are able to further reduce the error of matches that are already made in close proximity to the true match, disambiguating spatially close places. \textbf{Bottom:} For lower performing baselines, such as NetVLAD, our \textcolor{HOPS}{HOPS} fused descriptors corrected a high number of large errors as well.}
        \label{fig:error_hist}
    \end{figure}

    Figure~\ref{fig:error_hist} provides insights into how the \textcolor{HOPS}{HOPS} fused descriptors are improving VPR performance. It shows that they are, especially for already high performance baseline methods, further reducing the metric error of place matches that are already quite close to the ground truth match. This is a different phenomenon to typical improvements in VPR where egregiously wrong matches are ``corrected'' by improved features to fall within the correct zone around the ground truth. We suspect the reason for this is that the stacking/fusing of multiple reference descriptors for each place is reducing the volatility of matching in the region near the ground truth location (in datasets where subsequent frames often belong to a similar spatial location), meaning the true best match is less likely to be ``outmatched'' by a nearby visually similar images. For VPR descriptors with lower baseline performance, such as NetVLAD, there are still a high number of large errors corrected as well.

    \setlength\tabcolsep{0.5mm} %

\begin{table*}[!ht]
    \caption{Recall@1 on Nordland datasets: See Table~\ref{table:HDC_refs_RobotCar} for format conventions. Our \textcolor{HOPS}{HOPS} fused descriptor outperforms the best single-reference results in \textbf{23/24} cases and the other multi-reference approaches in \textbf{18/24} cases.}
    \centering
    \scriptsize
    \begin{tabular}{l*{4}{c}|*{4}{c}|*{4}{c}|*{4}{c}|*{4}{c}|*{4}{c}}
        \toprule
        \textbf{Queries $\rightarrow$} 
        & \rotatebox{90}{\raisebox{0.075cm}{\textbf{Fall}}} & \rotatebox{90}{\raisebox{0.075cm}{\textbf{Spring}}} & \rotatebox{90}{\raisebox{0.075cm}{\textbf{Summer}}} & \rotatebox{90}{\raisebox{0.075cm}{\textbf{Winter}}}  
        & \rotatebox{90}{\raisebox{0.075cm}{\textbf{Fall}}} & \rotatebox{90}{\raisebox{0.075cm}{\textbf{Spring}}} & \rotatebox{90}{\raisebox{0.075cm}{\textbf{Summer}}} & \rotatebox{90}{\raisebox{0.075cm}{\textbf{Winter}}}  
        & \rotatebox{90}{\raisebox{0.075cm}{\textbf{Fall}}} & \rotatebox{90}{\raisebox{0.075cm}{\textbf{Spring}}} & \rotatebox{90}{\raisebox{0.075cm}{\textbf{Summer}}} & \rotatebox{90}{\raisebox{0.075cm}{\textbf{Winter}}} 
        & \rotatebox{90}{\raisebox{0.075cm}{\textbf{Fall}}} & \rotatebox{90}{\raisebox{0.075cm}{\textbf{Spring}}} & \rotatebox{90}{\raisebox{0.075cm}{\textbf{Summer}}} & \rotatebox{90}{\raisebox{0.075cm}{\textbf{Winter}}} 
        & \rotatebox{90}{\raisebox{0.075cm}{\textbf{Fall}}} & \rotatebox{90}{\raisebox{0.075cm}{\textbf{Spring}}} & \rotatebox{90}{\raisebox{0.075cm}{\textbf{Summer}}} & \rotatebox{90}{\raisebox{0.075cm}{\textbf{Winter}}} 
        & \rotatebox{90}{\raisebox{0.075cm}{\textbf{Fall}}} & \rotatebox{90}{\raisebox{0.075cm}{\textbf{Spring}}} & \rotatebox{90}{\raisebox{0.075cm}{\textbf{Summer}}} & \rotatebox{90}{\raisebox{0.075cm}{\textbf{Winter}}} \\
        \hline
        \textbf{References} & \multicolumn{4}{c|}{\textbf{NetVLAD (4096D)}} & \multicolumn{4}{c|}{\textbf{SALAD (8448D)}} & \multicolumn{4}{c|}{\textbf{MixVPR (4096D)}} & \multicolumn{4}{c|}{\textbf{CosPlace (512D)}} & \multicolumn{4}{c|}{\textbf{EigenPlaces (512D)}} & \multicolumn{4}{c}{\textbf{CricaVPR (10752D)}}\\
        \hline
        Fall  & -                 & \underline{43.3}  & \underline{61.5} & 16.1  & 
        - & \underline{80.2} & \underline{79.9} & 72.8 &
        - & \underline{78.8} & \underline{78.8} & 66.9 & 
        - & \underline{76.9} & \underline{77.6} & 61.5 &
        - & \underline{77.5} & \underline{78.5} & 63.3 &
        - & \underline{81.6} & \underline{81.3} & 77.3 \\
        Spring                & 37.0              & -                 & 35.2            & \underline{16.2} &
        78.4 & - & 76.8 & \underline{75.8} & 
        73.3 & - & 69.3 & \underline{73.6} & 
        71.2 & - & 65.3 & \underline{70.5} &
        74.5 & - & 68.8 & \underline{67.5} &
        80.6 & - & 77.8 & \underline{79.6} \\
        Summer                & \underline{61.1}  & 41.0              & -               & 15.5        &
        \underline{80.0} & 78.2 & - & 71.1 &
        \underline{78.6} & 75.5 & - & 63.7 & 
        \underline{77.7} & 72.3 & - & 56.6 &
        \underline{78.8} & 74.3 & - & 59.5 &
        \underline{81.4} & 80.4 & - & 74.6 \\
        Winter                & 12.4              & 18.1              & 11.9            & -           &
        71.0 & 76.9 & 69.3 & - &
        57.2 & 70.9 & 52.9 & - & 
        51.2 & 68.4 & 46.5 & - &
        57.1 & 71.0 & 52.9 & - &
        73.9 & 79.8 & 70.9 & - \\
        \hline
        dMat Avg~\cite{fischer2020event}  & 57.3              & 56.8              & 55.7            &  \textbf{26.6}       &
        81.2 & 81.7 & 80.0 & 79.4 &
        80.3 & 81.3 & 77.9 & 76.5 & 
        77.7 & 79.6 & 73.8 & 70.9  &
        79.9 & 80.4 & 77.2 & 72.5 &
        83.2 & 83.2 & 81.5 & 82.1 \\
        Pooling  & 63.2              & 50.9              & 62.9            &  18.5       &
        81.5 & 81.9 & 80.5 & 77.3 &
        80.8 & 81.5 & \textbf{79.9} & 75.8 & 
        80.4 & 80.3 & \textbf{78.6} & \textbf{71.5} &
        81.0 & 81.1 & \textbf{79.3} & 68.7 &
        82.9 & \textbf{84.1} & \textbf{82.5} & 81.1 \\
        \textcolor{HOPS}{HOPS} (Ours)     & \textbf{63.5}     & \textbf{62.7}     & \textbf{63.3}   &  25.7       &
        \textbf{82.1} & \textbf{82.0} & \textbf{80.7} & \textbf{79.7} &
        \textbf{81.7} & \textbf{81.8} & 79.2 & \textbf{77.1} & 
        \textbf{80.4} & \textbf{81.2} & 76.6 & 71.3 &
        \textbf{81.2} & \textbf{81.6} & 78.6 & \textbf{72.7} &
        \textbf{83.9} & 83.8 & \textbf{82.5} & \textbf{82.4} \\
        \bottomrule
    \end{tabular}
    \label{table:HDC_refs_Nord}
\end{table*}
    \setlength\tabcolsep{0.5mm} %

\begin{table*}[!ht]
    \caption{Recall@1 on SFU-Mountain datasets: See Table~\ref{table:HDC_refs_RobotCar} for format conventions. Our \textcolor{HOPS}{HOPS} fused descriptor outperforms the best single-reference results in \textbf{100\%} of cases and the other multi-reference approaches in \textbf{29/36} cases.}
    \centering
    {\fontsize{5.5}{6.5}\selectfont
    \begin{tabular}{l*{6}{c}|*{6}{c}|*{6}{c}|*{6}{c}|*{6}{c}|*{6}{c}}
        \toprule
        \textbf{Queries $\rightarrow$} 
        & \rotatebox{90}{\raisebox{0.075cm}{\textbf{Dry}}} & \rotatebox{90}{\raisebox{0.075cm}{\textbf{Dusk}}} & \rotatebox{90}{\raisebox{0.075cm}{\textbf{Jan}}} & \rotatebox{90}{\raisebox{0.075cm}{\textbf{Nov}}} & \rotatebox{90}{\raisebox{0.075cm}{\textbf{Sept}}} & \rotatebox{90}{\raisebox{0.075cm}{\textbf{Wet}}}
        & \rotatebox{90}{\raisebox{0.075cm}{\textbf{Dry}}} & \rotatebox{90}{\raisebox{0.075cm}{\textbf{Dusk}}} & \rotatebox{90}{\raisebox{0.075cm}{\textbf{Jan}}} & \rotatebox{90}{\raisebox{0.075cm}{\textbf{Nov}}} & \rotatebox{90}{\raisebox{0.075cm}{\textbf{Sept}}} & \rotatebox{90}{\raisebox{0.075cm}{\textbf{Wet}}}
        & \rotatebox{90}{\raisebox{0.075cm}{\textbf{Dry}}} & \rotatebox{90}{\raisebox{0.075cm}{\textbf{Dusk}}} & \rotatebox{90}{\raisebox{0.075cm}{\textbf{Jan}}} & \rotatebox{90}{\raisebox{0.075cm}{\textbf{Nov}}} & \rotatebox{90}{\raisebox{0.075cm}{\textbf{Sept}}} & \rotatebox{90}{\raisebox{0.075cm}{\textbf{Wet}}}
        & \rotatebox{90}{\raisebox{0.075cm}{\textbf{Dry}}} & \rotatebox{90}{\raisebox{0.075cm}{\textbf{Dusk}}} & \rotatebox{90}{\raisebox{0.075cm}{\textbf{Jan}}} & \rotatebox{90}{\raisebox{0.075cm}{\textbf{Nov}}} & \rotatebox{90}{\raisebox{0.075cm}{\textbf{Sept}}} & \rotatebox{90}{\raisebox{0.075cm}{\textbf{Wet}}}
        & \rotatebox{90}{\raisebox{0.075cm}{\textbf{Dry}}} & \rotatebox{90}{\raisebox{0.075cm}{\textbf{Dusk}}} & \rotatebox{90}{\raisebox{0.075cm}{\textbf{Jan}}} & \rotatebox{90}{\raisebox{0.075cm}{\textbf{Nov}}} & \rotatebox{90}{\raisebox{0.075cm}{\textbf{Sept}}} & \rotatebox{90}{\raisebox{0.075cm}{\textbf{Wet}}}
        & \rotatebox{90}{\raisebox{0.075cm}{\textbf{Dry}}} & \rotatebox{90}{\raisebox{0.075cm}{\textbf{Dusk}}} & \rotatebox{90}{\raisebox{0.075cm}{\textbf{Jan}}} & \rotatebox{90}{\raisebox{0.075cm}{\textbf{Nov}}} & \rotatebox{90}{\raisebox{0.075cm}{\textbf{Sept}}} & \rotatebox{90}{\raisebox{0.075cm}{\textbf{Wet}}} \\
        \hline
        \textbf{References} & \multicolumn{6}{c|}{\textbf{NetVLAD (4096D)}} & \multicolumn{6}{c|}{\textbf{SALAD (8448D)}} & \multicolumn{6}{c|}{\textbf{MixVPR (4096D)}} & \multicolumn{6}{c|}{\textbf{CosPlace (512D)}} & \multicolumn{6}{c|}{\textbf{EigenPlaces (512D)}} & \multicolumn{6}{c}{\textbf{CricaVPR (10752D)}}\\
        \hline
        Dry                  & -                & 43.9             & 25.5             & 33.0             & 23.6            & 38.4     &
        -                & \underline{99.0} & 92.5             & \underline{96.9} & 94.8           & 96.6 &
        -                & 94.3             & 81.6              & 89.9             & 86.8             & 92.0   &
        - & 91.7 & 79.2 & 82.9 & 81.0 & 88.6 &
        - & 92.5 & 83.1 & 87.8 & 87.0 & 93.0 &
        - & 98.7 & 91.9 & 95.8 & 93.0 & 97.9 \\ 
        Dusk                 & \underline{52.7} & -                & \underline{28.6} & 36.9             & \underline{34.0} & \underline{62.1}     &
        \underline{99.0} & -                & \underline{95.6} & 96.1             & 94.0           & \underline{98.2} &
        \underline{98.4} & -                & 90.9              & 94.3             & \underline{93.3} & \underline{98.4}   &
        91.7 & - & 82.1 & 84.9 & 77.7 & \underline{94.8} &
        \underline{95.1} & - & 89.4 & 90.4 & 88.1 & \underline{97.1} &
        \underline{99.2} & - & \underline{97.4} & \underline{96.6} & 94.8 & \underline{99.0} \\ 
        Jan                  & 25.5             & 34.6             & -                & 26.5             & 21.8            & 31.2     &
        94.6             & 96.9             & -                & 95.8             & 93.5          & 93.5 &
        75.1             & 84.4             & -                 & 71.7             & 70.7             & 79.5   &
        81.0 & 86.5 & - & 77.9 & 70.6 & 85.5 &
        86.8 & 88.3 & - & 82.1 & 77.7 & 86.8 &
        95.6 & 96.4 & - & 94.5 & 93.5 & 95.6 \\ 
        Nov                  & 30.1             & 31.2             & 23.1             & -                & 33.8            & 32.7     &
        95.3             & 94.0             & 94.8             & -                & \underline{96.4} & 96.4 &
        86.0             & 84.2             & 75.6              & -                & 92.2             & 88.1   &
        80.8 & 80.8 & 72.5 & - & \underline{89.9} & 80.8 &
        88.3 & 86.0 & 79.5 & - & \underline{93.5} & 87.0 &
        94.5 & 96.1 & 93.8 & - & \underline{97.7} & 96.1 \\ 
        Sept                 & 27.0             & 30.7             & 20.5             & \underline{38.4} & -               & 29.1     &
        94.0             & 88.8             & 93.5             & 95.3             & -              & 92.5 &
        84.9             & 86.5             & 75.1              & 94.0             & -                & 85.5  &
        77.9 & 75.1 & 68.3 & \underline{89.6} & - & 75.8 &
        81.8 & 80.8 & 73.5 & \underline{92.7} & - & 83.9 &
        92.7 & 91.9 & 93.0 & 95.8 & - & 90.9 \\ 
        Wet                  & 44.4             & \underline{63.9} & 28.3             & 38.2             & 28.8            & -        &
        97.7             & 98.7 & 94.6             & 95.1             & 93.5           & -  &
        95.8             & \underline{96.9} & \underline{92.7}  & \underline{95.1} & 92.5             & -   &
        \underline{94.0} & \underline{95.1} & \underline{84.9} & 88.3 & 84.7 & - &
        \underline{95.1} & \underline{97.1} & \underline{91.7} & 91.2 & 89.9 & - &
        97.1 & \underline{99.2} & 96.6 & \underline{96.6} & 93.5 & - \\ 
        \hline
        dMat Avg~\cite{fischer2020event} & 63.4 & 62.3 & 40.5 & 61.0 & 48.3 & 66.8 &
        99.5 & 99.5 & 98.7 & 99.0 & 98.4 & 99.2 &
        99.2 & 98.4 & 93.2 & \textbf{99.5} & 97.4 & 98.7 &
        95.6 & 96.6 & 87.3 & 94.5 & 93.8 & 96.1 &
        97.1 & 98.2 & 94.0 & 98.2 & 96.1 & 97.9 &
        99.5 & 99.7 & \textbf{98.7} & 99.5 & 98.2 & 99.5 \\
        Pooling & 59.7             & 68.8             & 38.2             & 50.9             & 42.6            & 66.2     &
        \textbf{99.7}    & 99.2             & 98.2             & 98.0             & 97.1           & \textbf{99.5}  &
        99.2             & 98.4             & 95.3              & 98.7             & 96.6             & \textbf{99.7} &
        \textbf{98.4} & 98.2 & 93.2 & 96.1 & \textbf{95.8} & \textbf{98.4} &
        \textbf{99.0} & 98.4 & 95.3 & 95.6 & \textbf{96.4} & \textbf{99.0} &
        99.5 & 99.5 & 97.9 & 98.4 & \textbf{99.0} & 99.2 \\ 
        \textcolor{HOPS}{HOPS} (Ours)    & \textbf{68.3}    & \textbf{74.6}    & \textbf{47.5}    & \textbf{68.1}    & \textbf{56.6}   & \textbf{76.1}     &
        \textbf{99.7}    & \textbf{100}     & \textbf{99.2}    & \textbf{99.5}    & \textbf{98.7}  & 99.2  &
        \textbf{99.5}    & \textbf{99.5}    & \textbf{97.1}     & \textbf{99.5}    & \textbf{97.9}    & 99.5   &
        97.9 & \textbf{98.7} & \textbf{95.1} & \textbf{96.9} & 95.6 & 98.2 &
        98.7 & \textbf{99.5} & \textbf{97.4} & \textbf{99.2} & 96.1 & \textbf{99.0} &
        \textbf{100.0} & \textbf{100.0} & \textbf{98.7} & \textbf{99.7} & \textbf{99.0} & \textbf{99.7} \\
        \bottomrule
    \end{tabular}
    }
    \label{table:HDC_refs_SFU}
    \vspace{-\baselineskip}
\end{table*}

    \subsection{Comparisons to Multi Reference Set Baselines}
    \label{subsec:multi-ref_results}
    With respect to the multi-reference set approaches, \mbox{Tables~\ref{table:HDC_refs_RobotCar}-\ref{table:HDC_refs_SFU}} show that while the distance matrix averaging and pooling methods typically provide improvements over single-reference methods, the HDC fused descriptors provide the highest R@1 in $69$ out of $90$ cases. %
    In addition, we reiterate that the HDC fused descriptors maintain the same computation and memory costs as in the single-reference set approaches which provides significant advantage over the pooling and averaging approaches whose computational and storage complexities increase linearly with the number of reference sets.
    
    One can observe that the reference pooling approach is more performant for lower dimensional descriptors such as CosPlace and EigenPlaces, whereas the distance matrix averaging performs better for the other higher dimensional descriptors –– as highlighted in the previous subsection, these results are not surprising given that HDC assumes high-dimensional feature vectors but both CosPlace and EigenPlaces are relatively low-dimensional.

    \subsection{Reducing Dimensionality}
    \label{subsec:red-dim_results}
    For large scale image retrieval tasks, the size of image descriptors can have a significant effect on the computational overhead and required memory allocation. Here, we investigate the possible advantages \textcolor{HOPS}{HOPS} fused descriptors have for reducing the dimensionality of existing SOTA VPR methods. That is, given a VPR descriptor and a selection of separate reference sets which achieve a certain performance, how can \textcolor{HOPS}{HOPS} fused descriptors reduce dimensionality while still matching or exceeding this original performance. For these experiments, we used a Gaussian Random Projection to reduce descriptor dimensionality for its relevance being based in high dimensional spaces (Section~\ref{subsec::GRP}), however, this method could be substituted with other dimensionality reduction methods.

    \begin{figure}[h!t]
        \centering
        \setlength\tabcolsep{1.2mm}
        \begin{tabular}{cc}
            \small{\textbf{CosPlace (512D)}} & \small{\textbf{MixVPR (4096D)}} \\
            \includegraphics[width=0.5375\linewidth]{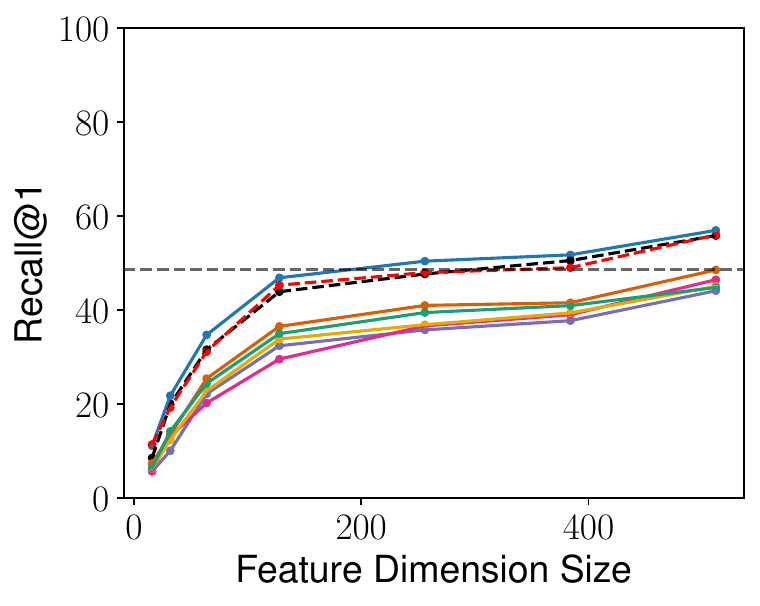} &  
            \includegraphics[width=0.4625\linewidth]{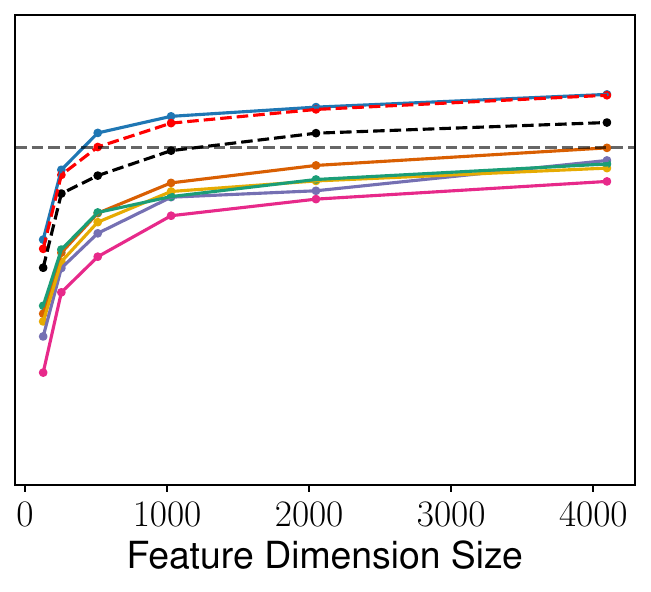} \\
            \small{\textbf{SALAD (8448D)}} & \small{\textbf{CricaVPR (10752D)}} \\
            \includegraphics[width=0.5375\linewidth]{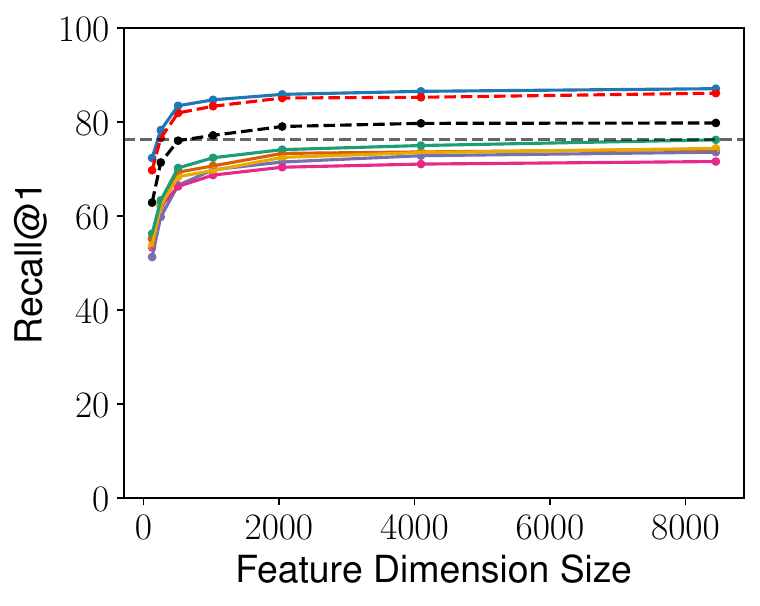} & 
            \includegraphics[width=0.4625\linewidth]{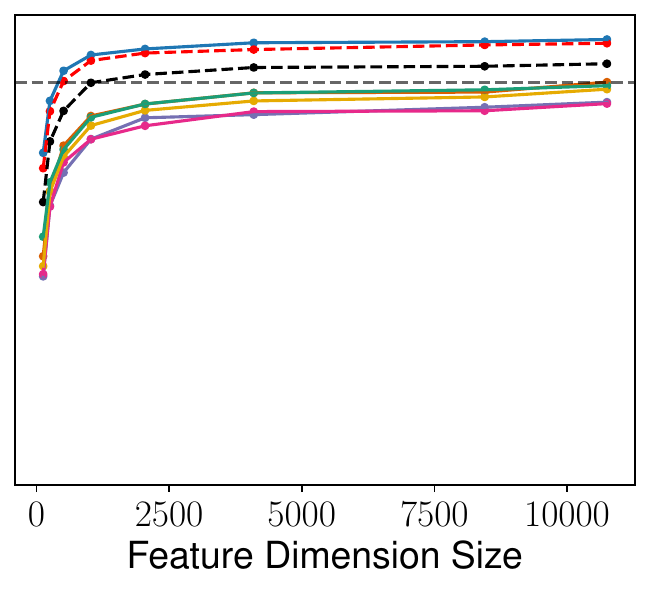} \\
            \multicolumn{2}{r}{\includegraphics[width=\linewidth]{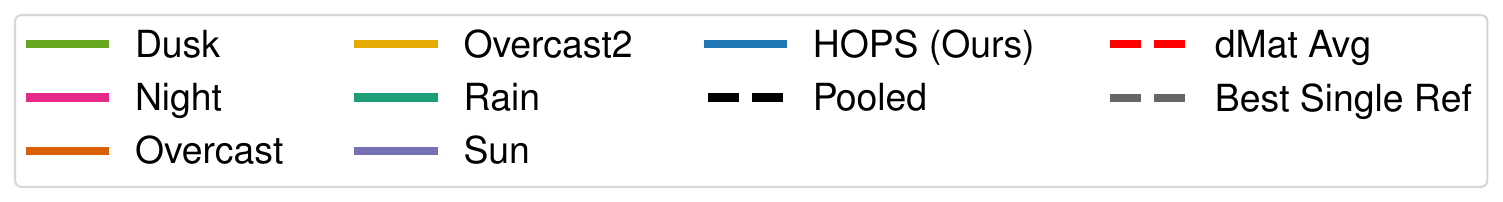}}
        \end{tabular}
        \caption{Recall@1 performance for different VPR descriptors across the Oxford RobotCar Dusk set as dimensionality is reduced using Gaussian Random Projection. Our \textcolor{HOPS}{HOPS} fused descriptors are able to maintain the highest R@1, allowing for an alternate use where descriptor dimensionality can be reduced by up to $97\%$ while exceeding the best single-reference performance at full-size.}
        \label{fig:ftr_dims_abl}
    \end{figure}
    
    Figure~\ref{fig:ftr_dims_abl} shows representative results using CosPlace, MixVPR, SALAD, and CricaVPR on the RobotCar Dusk dataset (see the Supplementary Material for full results). Our proposed \textcolor{HOPS}{HOPS} fused descriptors exceed the performance of the best full-sized single-reference results with a much smaller descriptor size; about a $50\%$ and $95\%$ reduction for CosPlace and CricaVPR, respectively, \ie a recall of $85.7\%$ for CricaVPR can either be obtained using the 10752D original descriptor or our 512D reduced-dimension fused \textcolor{HOPS}{HOPS} descriptor. Our \textcolor{HOPS}{HOPS} fused approach and single-reference approaches follow a similar trend, with performance gradually being more effected by dimensionality reduction before a sudden drop off in R@1 -- importantly, our \textcolor{HOPS}{HOPS} descriptors maintain the highest R@1 values across all descriptor dimensions.

    \subsection{Substituting Synthetic Image Augmentations}
    \label{subsec:synth_results}
    So far, we have explored multi-reference VPR approaches with the assumption that multiple reference sets have been collected from real-world data. However, here we show that multiple reference sets can also be created by synthetically augmenting a single reference dataset. This is one possible way to enable the use of our \textcolor{HOPS}{HOPS} fused descriptors in single-reference scenarios.

    Table~\ref{table:RobotCar_Synth} shows a proof of concept study where image augmentations such as synthetic darkening of an image (generated using~\cite{luo2023similarity}), the application of Poisson noise, and downsampling and re-upsampling an image are used to exploit some of the performance benefits of \textcolor{HOPS}{HOPS} fused descriptors without requiring \textit{real} multiple reference traverses.
    
    For the RobotCar Dusk and Night sets, \textcolor{HOPS}{HOPS} fused descriptors using the synthetic condition changes improve R@1 by absolute $2.5\%$ and $2.2\%$ respectively over the best single-reference results.
    We note that while we improve performance on average by 1.0\%, in the Overcast query the performance reduces slightly by 0.6\%.
    
    \begin{table}[t]
    \caption{Recall@1 on RobotCar datasets Using Synthetic Changes}
    \centering
    \setlength\tabcolsep{1.2mm}
    \resizebox{\linewidth}{!}{
        \begin{tabular}{|c|ccccc|}
            \hline
            \textbf{Queries $\rightarrow$} & \multicolumn{5}{c|}{\textbf{DinoV2 SALAD (8448D)}} \\
            \cline{2-6}
            \textbf{References $\downarrow$} & \textbf{Dusk}         & \textbf{Night}        & \textbf{Overcast}     & \textbf{Overcast2}     & \textbf{Rain} \\
            \hline
            \multicolumn{1}{|c|}{Sunny} & \underline{73.6} & \underline{70.5} & \underline{84.8} & \underline{88.6} & \underline{87.3} \\
            \multicolumn{1}{|c|}{Synthetic Dark~\cite{luo2023similarity}} & 70.9 & 68.4 & 73.2 & 77.7 & 77.4 \\
            \multicolumn{1}{|c|}{Poisson Noise} & 64.3 & 60.6 & 77.1 & 80.8 & 79.9 \\
            \multicolumn{1}{|c|}{Downsample-Upsample} & 68.8 & 67.0 & 80.1 & 83.2 & 82.7 \\
            \hline
            \multicolumn{1}{|c|}{dMat Avg~\cite{fischer2020event}} & 75.8 & \textbf{73.2} & 82.8 & 87.3 & 86.5 \\
            \multicolumn{1}{|c|}{Pooling} & 73.5 & 69.4 & \textbf{84.5} & 88.5 & 86.8 \\
            \multicolumn{1}{|c|}{\textcolor{HOPS}{HOPS} (Ours)} & \textbf{76.1} & 72.7 & 84.2 & \textbf{88.8} & \textbf{87.7} \\
            \hline
        \end{tabular}
    }
    \label{table:RobotCar_Synth}
    \vspace{-\baselineskip}
\end{table}

    \subsection{Dataset Identification}
    \label{subsec:other_apps}
    Here we provide a brief investigation into another possible application of descriptor fusing via hyperdimensional computing: identifying in which environment one is located based on a single descriptor, \ie all reference descriptors of a dataset are fused into a \emph{single overall dataset descriptor}. Individual query descriptors from each of the datasets (and not from any reference set), can then be compared to these dataset descriptors to determine which dataset the query is from. By using all available non-query sets for each dataset and fusing them, this results in dataset identification with an accuracy $>\!99.7\%$ for all datasets. Full details can be found in the supplemental material.

    \section{Conclusion}
\label{sec::conclusion}
This paper investigated how reference sets captured under varying conditions can be fused with minimal compute and storage overhead using a hyperdimensional computing framework, to improve VPR performance under appearance change. Through an extensive set of experiments, we demonstrated that our \textcolor{HOPS}{HOPS} fused descriptors improve recall@1 over the best single-reference results for several multi-condition datasets and SOTA VPR methods. We also showed that while other multi-reference approaches also improve over the single-reference case, our \textcolor{HOPS}{HOPS} fused descriptors are generally the highest performing whilst also avoiding the computation and memory costs incurred in these other multi-reference approaches. This research further highlights the potential of the HDC framework for improving VPR, which is complementary to ongoing research efforts on extracting more invariant place features.

Multiple reference sets can be obtained both from real world sensory data but also from synthetically generated image transformations, especially when multiple reference sets are not available: we demonstrated the performance achievement of the latter when fusing descriptors from multiple image augmentations of a single reference set.

Finally this research also explored how a system can reduce computation and memory costs for real-time deployment without sacrificing performance: \textcolor{HOPS}{HOPS} fused descriptors can maintain the same performance as the best single-reference results whilst reducing the descriptor dimensionality by up to an order of magnitude. We also demonstrated how the HDC framework can be used to create whole dataset descriptors which can be used for identifying which dataset a query is from.

Future work can further improve both the capability and efficiency of HOPS descriptors by deeper investigating the effect of bundling on features and by exploring whether HOPS fused descriptors can be used to train more robust feature extractors. The work here primarily investigated the combination of multiple reference images from the \textit{same} location: preliminary investigation has also indicated that it is possible to stack together reference imagery from completely different datasets with no computational and minimal performance penalty, providing the possibility for highly compressible encoding of many maps into a single representation.

    {
        \small
        \bibliographystyle{ieeenat_fullname}
        \bibliography{main}
    }

    \clearpage
\setcounter{page}{1}
\maketitlesupplementary

\section{Overview}
In the following document, we provide supplemental material containing extended results for the experiments presented in the main manuscript. We begin by providing recall@1 results with the AnyLoc~\cite{keetha2023anyloc} Visual Place Recognition (VPR) descriptor in Section~\ref{sec:anyloc}, follow with additional figures showing dimensionality reduction performance on more query sets in Section~\ref{sec:supp_dim_red}, expand on results for our dataset identification experiment in Section~\ref{sec:supp_dset_id}, and finish with some qualitative VPR results in Section~\ref{sec:qual_results}.

\section{AnyLoc Recall@1}
\label{sec:anyloc}
Core results in the main manuscript (Sections~\ref{subsec:single-ref_results} and \ref{subsec:multi-ref_results}) demonstrate how our \textcolor{HOPS}{HOPS} fused descriptors can improve the recall@1 performance of multiple state-of-the-art (SOTA) VPR descriptors across a range of adverse conditions. Here, we present the recall@1 results for another recent SOTA VPR descriptor, AnyLoc~\cite{keetha2023anyloc}. We again use the implementation provided in the VPR method evaluation repository released with EigenPlaces\footnote{https://github.com/gmberton/VPR-methods-evaluation}, based on the original implementation and using the author-released weights. We note that this implementation does not include the PCA component for dimensionality reduction presented by the authors, and we could not find released weights for PCA.

Tables~\ref{table:anyloc_RC}--\ref{table:anyloc_SFU} show VPR performance for AnyLoc using single reference sets, multi-reference set approaches, and our \textcolor{HOPS}{HOPS} fused descriptor approach across all datasets used in the main manuscript. The tables demonstrate that the improvements to results observed for other SOTA VPR descriptors hold for AnyLoc as well, improving the recall@1 over the best single reference set recalls by at least absolute $7.7\%$, $4.6\%$, and $7.2\%$, and up to $14.3\%$, $14.6\%$, and $21\%$, respectively for the Oxford RobotCar~\cite{RobotCarDatasetIJRR}, Nordland~\cite{Niko13}, and SFU Mountain datasets~\cite{Jake2015SFU}.

For these AnyLoc results, our \textcolor{HOPS}{HOPS} fused descriptors achieve higher recall@1 than \textbf{both} the best single reference set \textbf{and} the other multi-reference set approaches in \textbf{all} cases. AnyLoc could be particularly suited to the use of hyperdimensional computing frameworks due to the large dimensionality of its feature vectors (49152D) prior to any dimensionality reduction.

\setlength\tabcolsep{1.5mm} %

\begin{table}[!ht]
    \caption{Recall@1 on the Oxford RobotCar dataset. }
    \centering
    \small
    \begin{tabular}{l*{5}{c}}
        \toprule
        \textbf{Queries $\rightarrow$} 
        & \rotatebox{90}{\raisebox{0.075cm}{\textbf{Dusk}}} & \rotatebox{90}{\raisebox{0.075cm}{\textbf{Night}}} & \rotatebox{90}{\raisebox{0.075cm}{\textbf{Overcast}}} & \rotatebox{90}{\raisebox{0.075cm}{\textbf{Overcast2}}} & \rotatebox{90}{\raisebox{0.075cm}{\textbf{Rain}}} \\
        \hline
        \textbf{References} & \multicolumn{5}{c}{\textbf{AnyLoc (49152D)}} \\
        \hline
        Sunny        & 63.1 & 60.6 & 75.3 & 82.9 & 80.6  \\ 
        Dusk      & - & \underline{68.7} & 48.5 & 52.3 & 57.4  \\ 
        Night      & \underline{69.7} & - & 49.4 & 49.4 & 51.7  \\ 
        Overcast      & 67.5 & 64.9 & - & \underline{83.0} & \underline{81.1}  \\
        Overcast2      & 65.6 & 61.7 & \underline{79.2} & - & 80.9  \\ 
        Rain      & 66.9 & 58.7 & 74.0 & 78.0 & -  \\ 

        \hline
        dMat Avg~\cite{fischer2020event} & 81.0 & 77.1 & 80.6 & 86.3 & 85.8  \\ 
        Pooling                          & 73.9 & 68.7 & 83.1 & 87.5 & 87.1  \\ 
        \textcolor{HOPS}{HOPS} (Ours)    & \textbf{84.0} & \textbf{79.0} & \textbf{86.9} & \textbf{90.7} & \textbf{91.4}  \\ 
        \bottomrule
    \end{tabular}
    \label{table:anyloc_RC}
    \vspace{\baselineskip}
\end{table}

\begin{table}[!ht]
    \caption{Recall@1 on the Nordland dataset. }
    \centering
    \small
    \begin{tabular}{l*{4}{c}}
        \toprule
        \textbf{Queries $\rightarrow$} 
        & \rotatebox{90}{\raisebox{0.075cm}{\textbf{Fall}}} & \rotatebox{90}{\raisebox{0.075cm}{\textbf{Spring}}} & \rotatebox{90}{\raisebox{0.075cm}{\textbf{Summer}}} & \rotatebox{90}{\raisebox{0.075cm}{\textbf{Winter}}} \\
        \hline
        \textbf{References} & \multicolumn{4}{c}{\textbf{AnyLoc (49152D)}} \\
        \hline
        Fall        & - & \underline{62.0} & \underline{70.5} & \underline{37.1} \\ 
        Spring      & 59.9 & - & 56.8 & 32.7 \\ 
        Summer      & \underline{70.9} & 57.1 & - & 33.0  \\ 
        Winter      & 24.7 & 36.2 & 22.4 & -  \\ 

        \hline
        dMat Avg~\cite{fischer2020event} & 73.5 & 71.1 & 71.4 & 47.9  \\ 
        Pooling                          & 73.4 & 66.9 & 71.2 & 35.5 \\ 
        \textcolor{HOPS}{HOPS} (Ours)    & \textbf{78.0} & \textbf{76.5} & \textbf{75.1} & \textbf{48.1} \\ 
        \bottomrule
    \end{tabular}
    \label{table:anyloc_Nord}
    \vspace{\baselineskip}
\end{table}

\begin{table}[!ht]
    \caption{Recall@1 on the SFU-Mountain dataset. }
    \centering
    \small
    \begin{tabular}{l*{6}{c}}
        \toprule
        \textbf{Queries $\rightarrow$} 
        & \rotatebox{90}{\raisebox{0.075cm}{\textbf{Dry}}} & \rotatebox{90}{\raisebox{0.075cm}{\textbf{Dusk}}} & \rotatebox{90}{\raisebox{0.075cm}{\textbf{Jan}}} & \rotatebox{90}{\raisebox{0.075cm}{\textbf{Nov}}} & \rotatebox{90}{\raisebox{0.075cm}{\textbf{Sept}}} & \rotatebox{90}{\raisebox{0.075cm}{\textbf{Wet}}} \\
        \hline
        \textbf{References} & \multicolumn{6}{c}{\textbf{AnyLoc (49152D)}} \\
        \hline
        Dry        & - & 77.7 & 57.4 & 66.0 & 62.1 & 68.1 \\ 
        Dusk      & \underline{84.7} & - & \underline{72.5} & 73.5 & 66.8 & \underline{89.4}\\ 
        Jan      & 56.1 & 69.6 & - & 60.0 & 52.7 & 70.1\\ 
        Nov      & 68.3 & 62.9 & 62.3 & - & \underline{71.7} & 65.7 \\ 
        Sept      & 61.3 & 59.7 & 56.6 & 71.7 & - & 62.6 \\ 
        Wet      & 77.7 & \underline{91.2} & 70.9 & \underline{77.4} & 63.9 & - \\ 

        \hline
        dMat Avg~\cite{fischer2020event} & 91.4 & 95.8 & 92.5 & 92.2 & 87.3 & 93.5 \\ 
        Pooling                          & 88.1 & 93.2 & 82.6 & 82.9 & 79.2 & 91.2 \\ 
        \textcolor{HOPS}{HOPS} (Ours)    & \textbf{97.4} & \textbf{98.4} & \textbf{93.5} & \textbf{97.1} & \textbf{92.5} & \textbf{97.1} \\ 
        \bottomrule
    \end{tabular}
    \label{table:anyloc_SFU}
\end{table}

\section{Dimensionality Reduction}
\label{sec:supp_dim_red}
\begin{figure}[!hb]
    \centering
    \setlength\tabcolsep{1.2mm}
    \begin{tabular}{cc}
        \multicolumn{2}{c}{\textbf{Query: RobotCar Night}} \\
        \small{\textbf{CosPlace (512D)}} & \small{\textbf{MixVPR (4096D)}} \\
        \includegraphics[width=0.5375\linewidth]{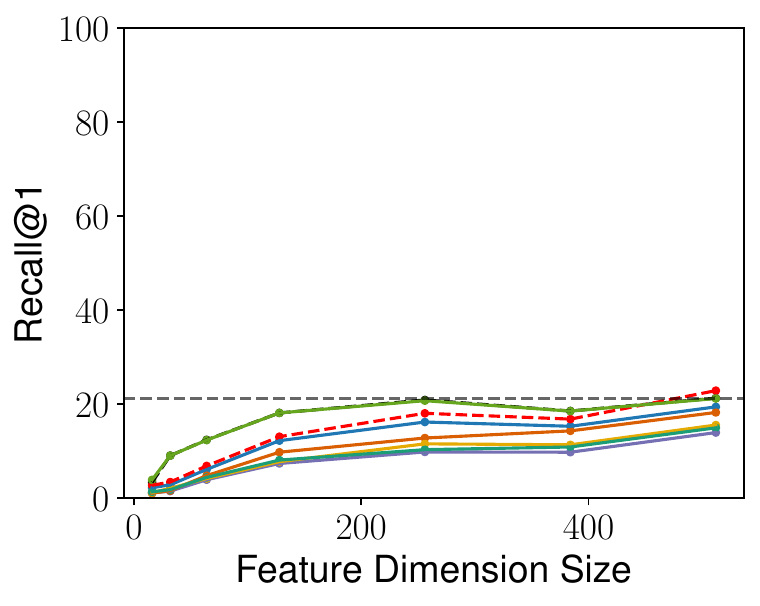} &  
        \includegraphics[width=0.4625\linewidth]{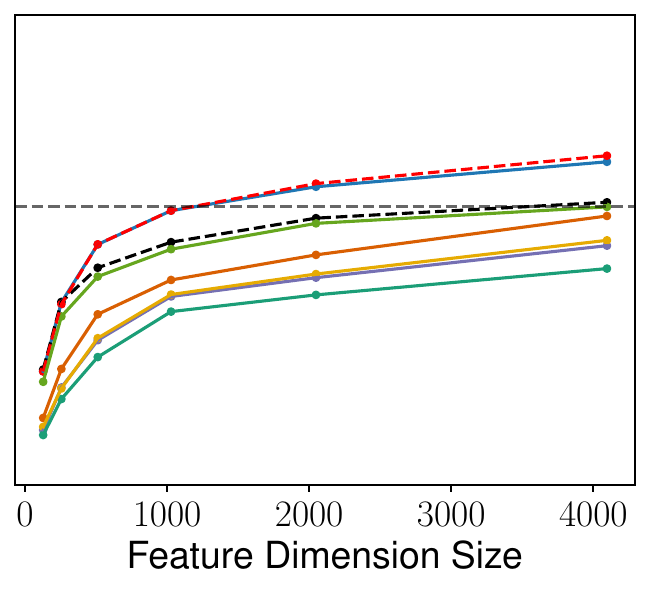} \\
        \small{\textbf{SALAD (8448D)}} & \small{\textbf{CricaVPR (10752D)}} \\
        \includegraphics[width=0.5375\linewidth]{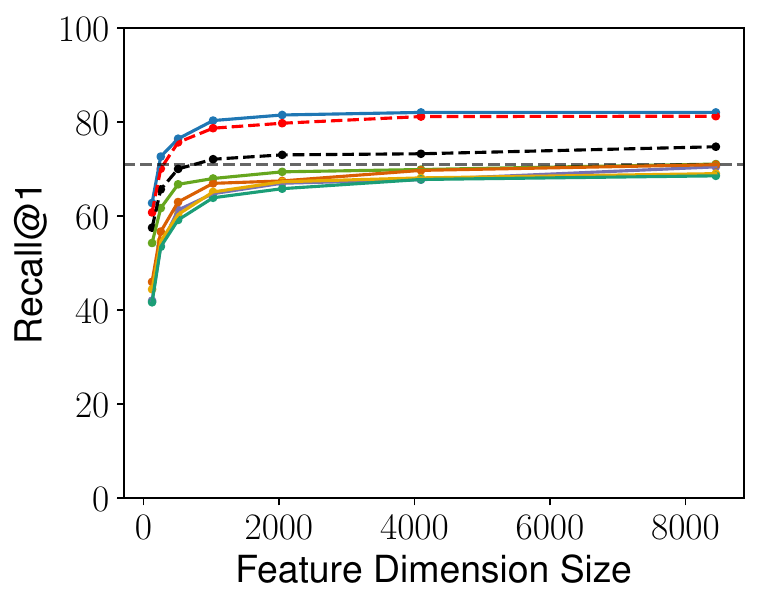} &
        \includegraphics[width=0.4625\linewidth]{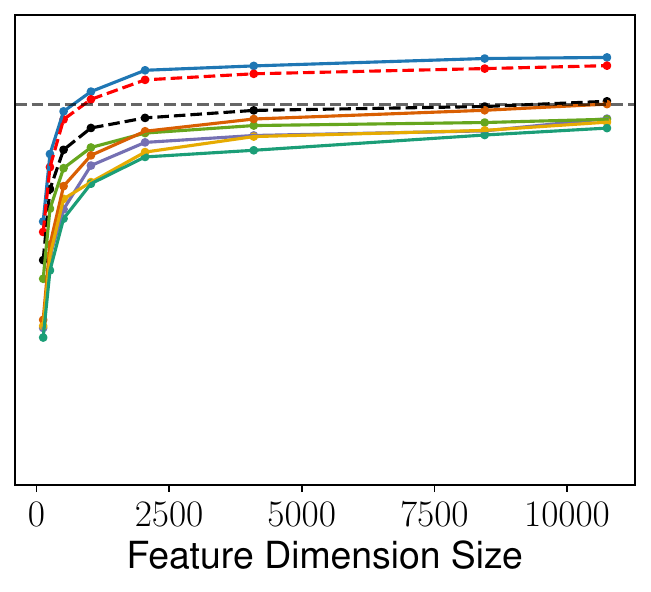} \\
        \small{\textbf{EigenPlaces (512D)}} & \small{\textbf{NetVLAD (4096D)}} \\
        \includegraphics[width=0.5375\linewidth]{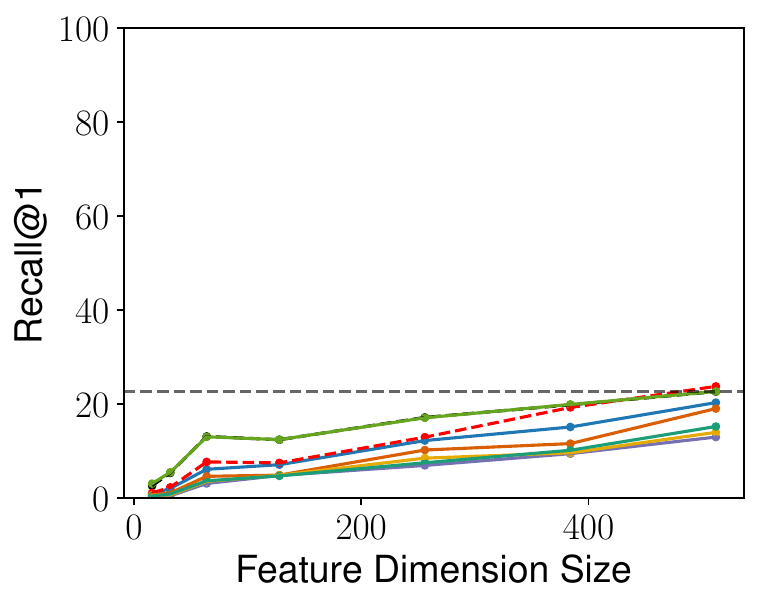} & 
        \includegraphics[width=0.4625\linewidth]{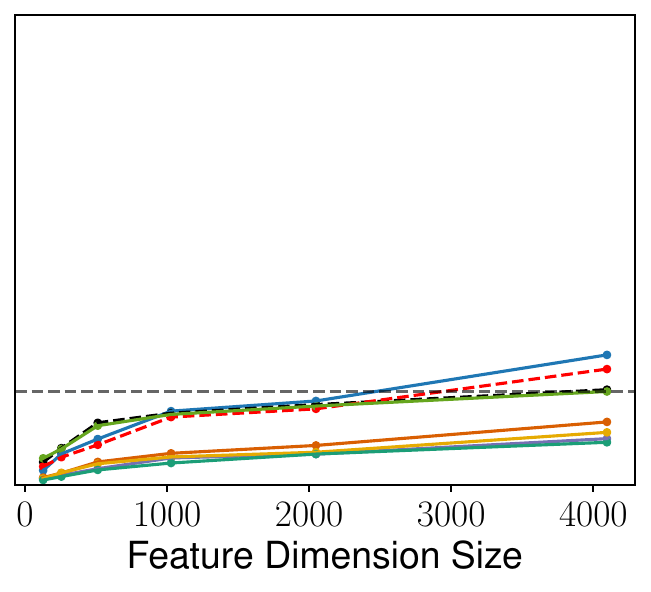} \\
        \multicolumn{2}{r}{\includegraphics[width=\linewidth]{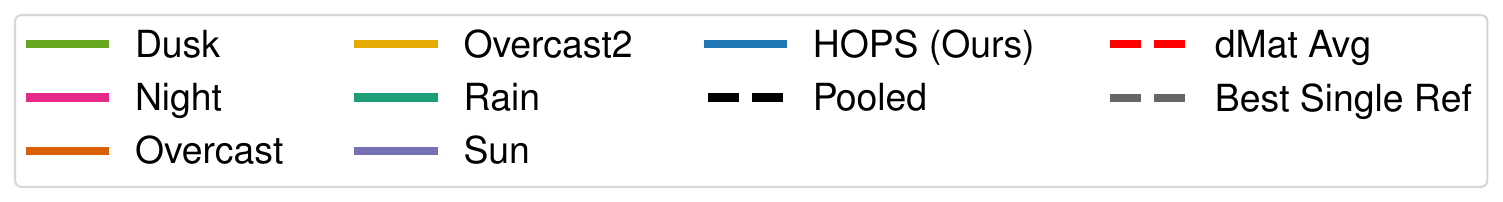}}
    \end{tabular}
    \caption{Recall@1 performance for different VPR descriptors across the Oxford RobotCar Night set as dimensionality is reduced using Gaussian Random Projection.}
    \label{fig:rc_night_ftr_red}
    \vspace{-\baselineskip}
\end{figure}

\begin{figure}[!hb]
    \centering
    \vspace{17.85\baselineskip}
    \setlength\tabcolsep{1.2mm}
    \begin{tabular}{cc}
        \multicolumn{2}{c}{\textbf{Query: RobotCar Rain}} \\
        \small{\textbf{CosPlace (512D)}} & \small{\textbf{MixVPR (4096D)}} \\
        \includegraphics[width=0.5375\linewidth]{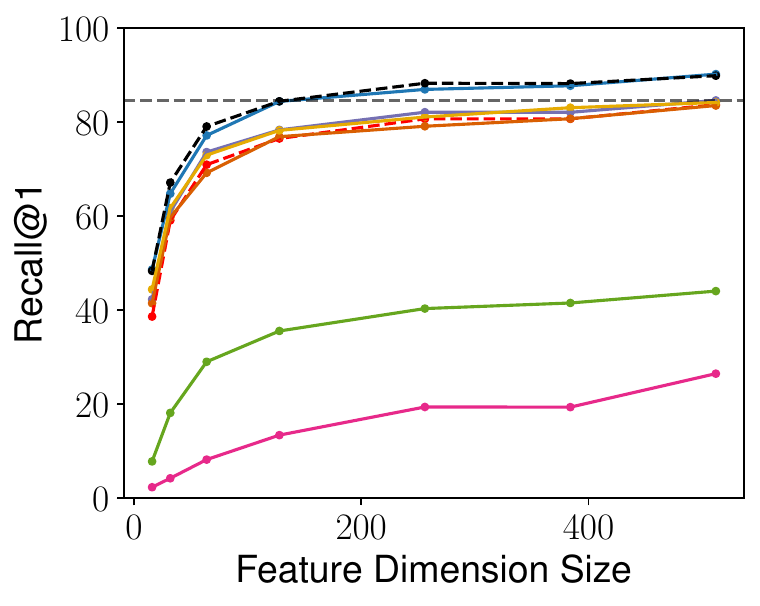} &  
        \includegraphics[width=0.4625\linewidth]{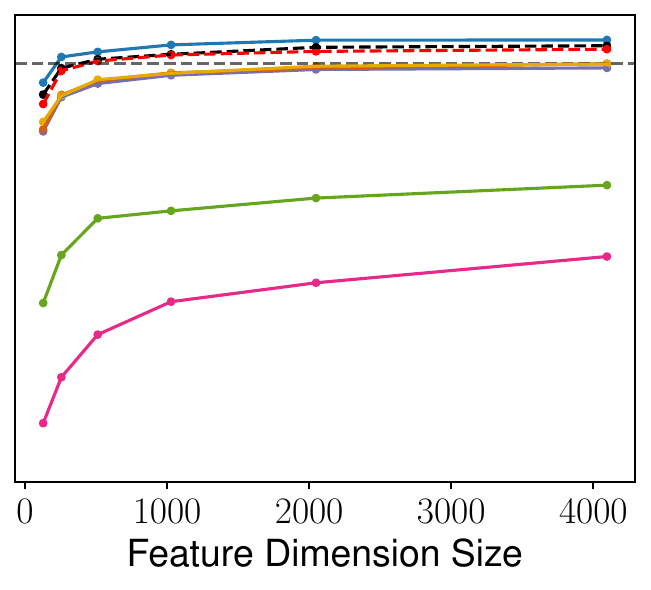} \\
        \small{\textbf{SALAD (8448D)}} & \small{\textbf{CricaVPR (10752D)}} \\
        \includegraphics[width=0.5375\linewidth]{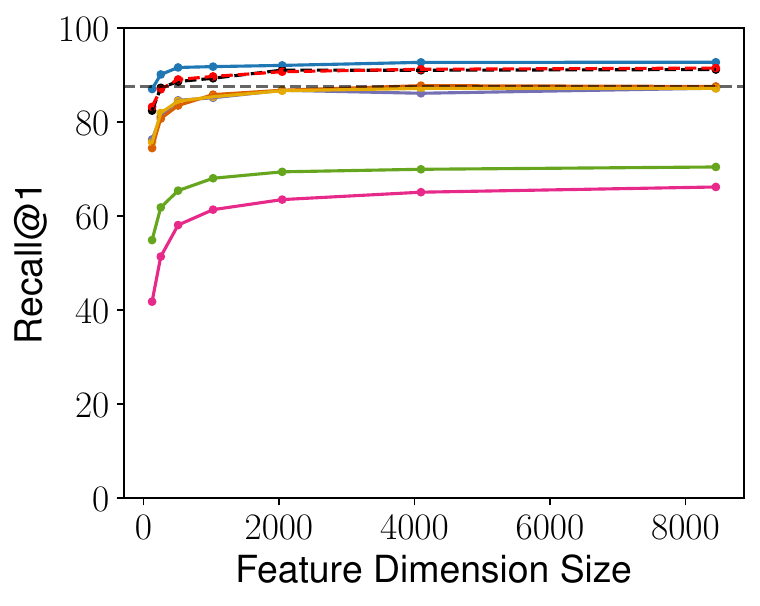} & 
        \includegraphics[width=0.4625\linewidth]{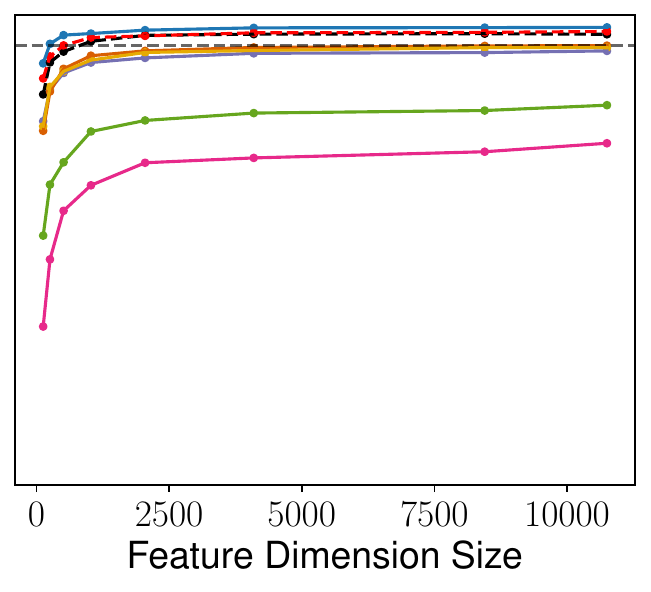} \\
        \small{\textbf{EigenPlaces (512D)}} & \small{\textbf{NetVLAD (4096D)}} \\
        \includegraphics[width=0.5375\linewidth]{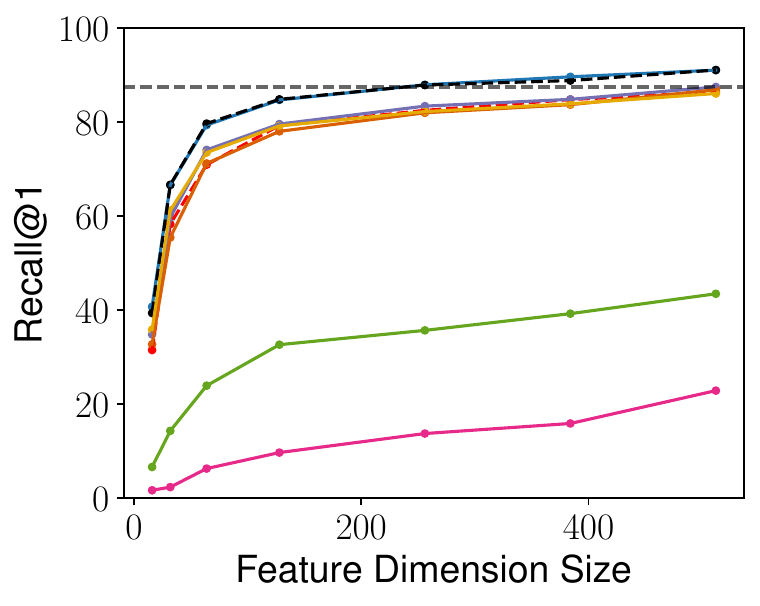} & 
        \includegraphics[width=0.4625\linewidth]{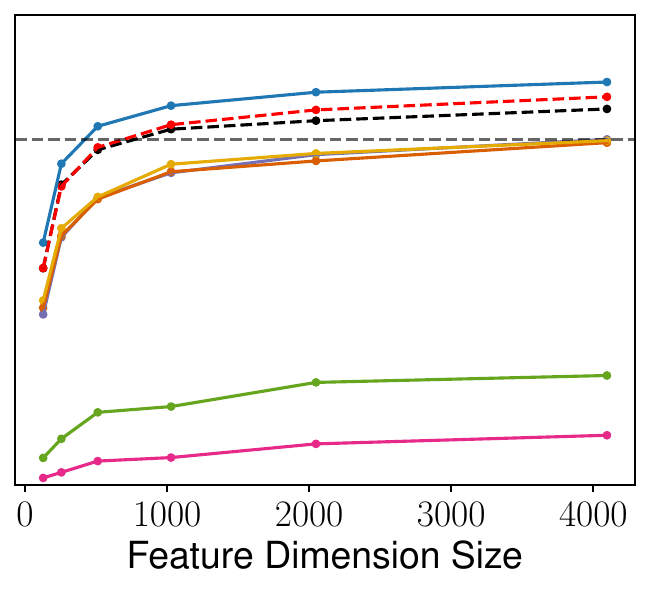} \\
        \multicolumn{2}{r}{\includegraphics[width=\linewidth]{figures/supp/ftr_reduction/RobotCar/ftr_red_abl_legend_v5.pdf}}
    \end{tabular}
    \caption{Recall@1 performance for different VPR descriptors across the Oxford RobotCar Rain set as dimensionality is reduced using Gaussian Random Projection.}
    \label{fig:rc_rain_ftr_red}
\end{figure}

\begin{figure}[h!t]
    \centering
    \setlength\tabcolsep{1.2mm}
    \begin{tabular}{cc}
        \multicolumn{2}{c}{\textbf{Query: Nordland Fall}} \\
        \small{\textbf{CosPlace (512D)}} & \small{\textbf{MixVPR (4096D)}} \\
        \includegraphics[width=0.5375\linewidth]{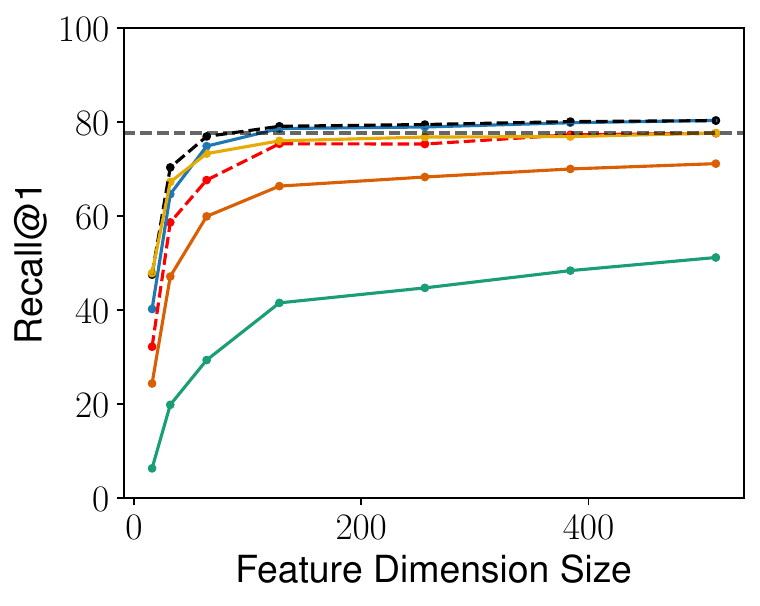} &  
        \includegraphics[width=0.4625\linewidth]{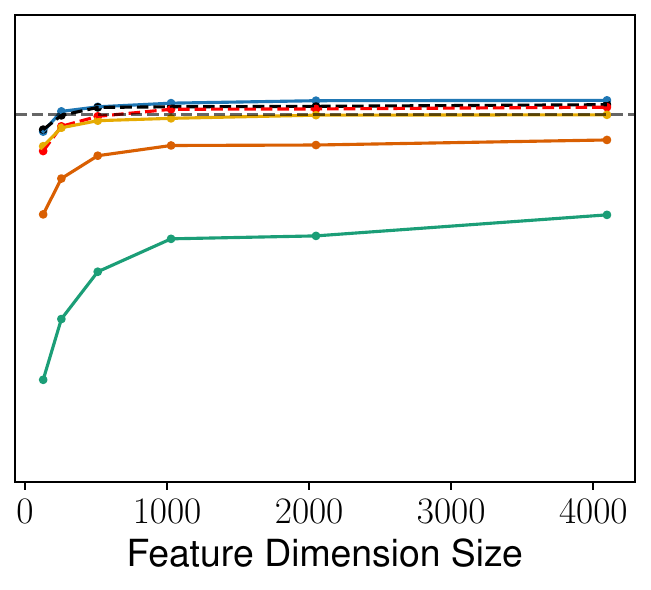} \\
        \small{\textbf{SALAD (8448D)}} & \small{\textbf{CricaVPR (10752D)}} \\
        \includegraphics[width=0.5375\linewidth]{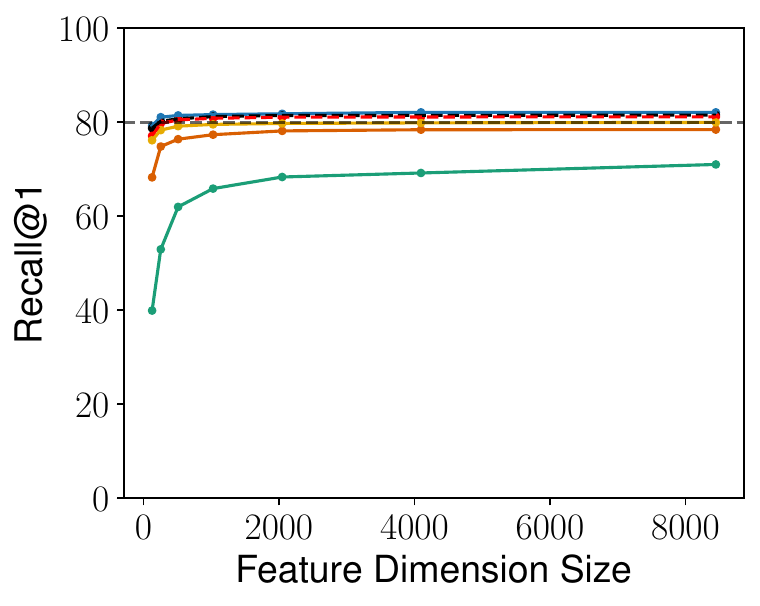} & 
        \includegraphics[width=0.4625\linewidth]{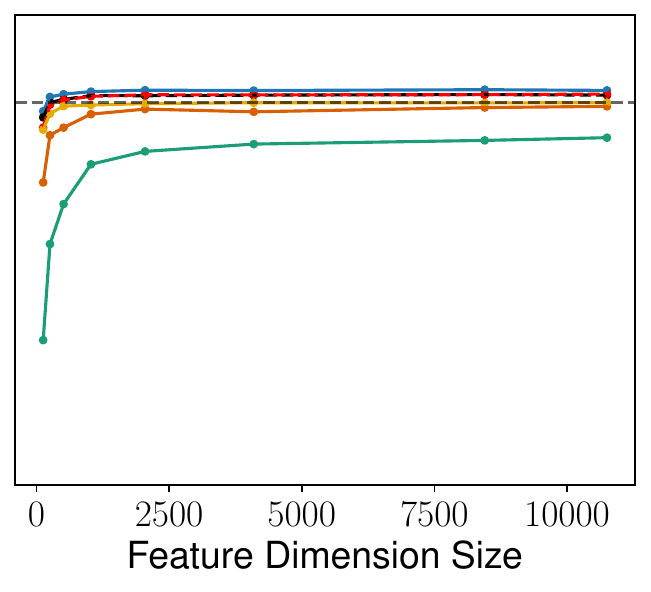} \\
        \small{\textbf{EigenPlaces (512D)}} & \small{\textbf{NetVLAD (4096D)}} \\
        \includegraphics[width=0.5375\linewidth]{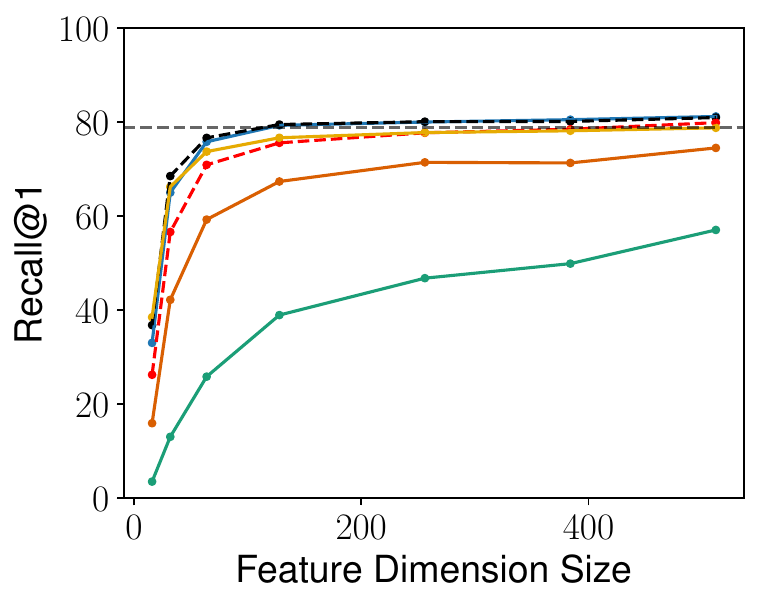} & 
        \includegraphics[width=0.4625\linewidth]{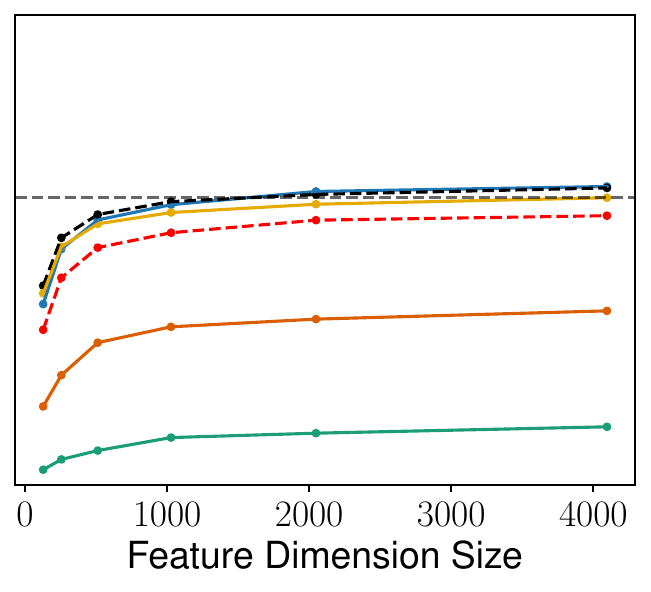} \\
        \multicolumn{2}{r}{\includegraphics[width=\linewidth]{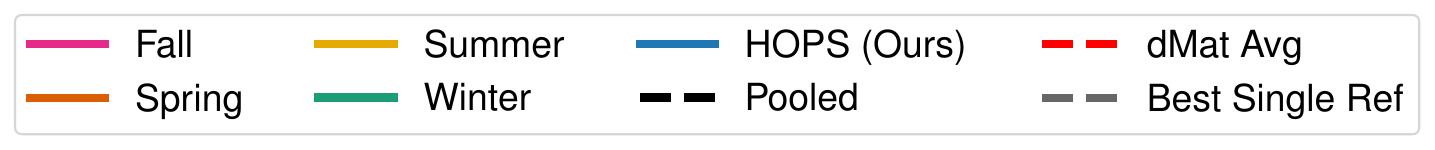}}
    \end{tabular}
    \caption{Recall@1 performance for different VPR descriptors across the Nordland Fall set as dimensionality is reduced using Gaussian Random Projection.}
    \label{fig:nord_fall_ftr_red}
\end{figure}

\begin{figure}[h!t]
    \centering
    \setlength\tabcolsep{1.2mm}
    \begin{tabular}{cc}
        \multicolumn{2}{c}{\textbf{Query: Nordland Winter}} \\
        \small{\textbf{CosPlace (512D)}} & \small{\textbf{MixVPR (4096D)}} \\
        \includegraphics[width=0.5375\linewidth]{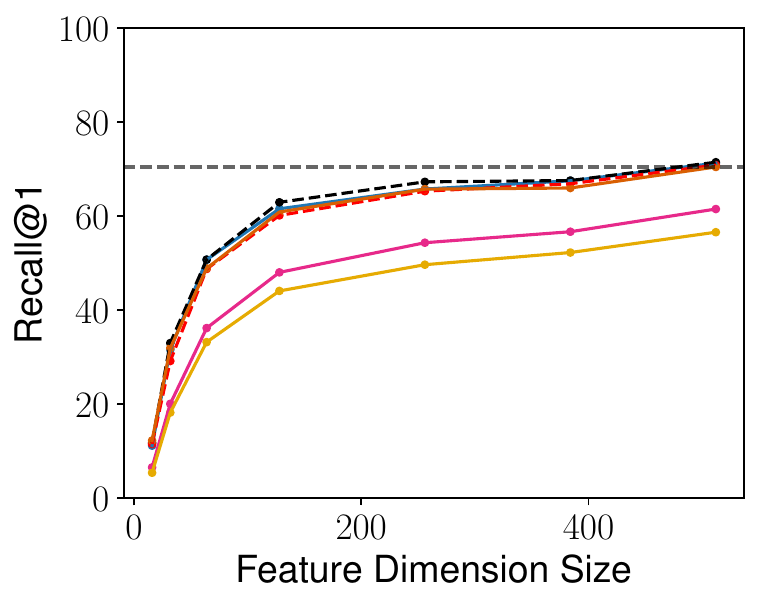} &  
        \includegraphics[width=0.4625\linewidth]{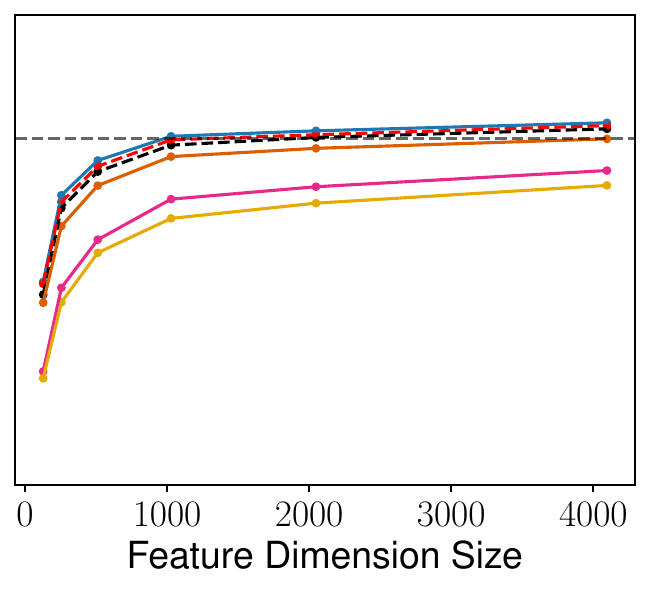} \\
        \small{\textbf{SALAD (8448D)}} & \small{\textbf{CricaVPR (10752D)}} \\
        \includegraphics[width=0.5375\linewidth]{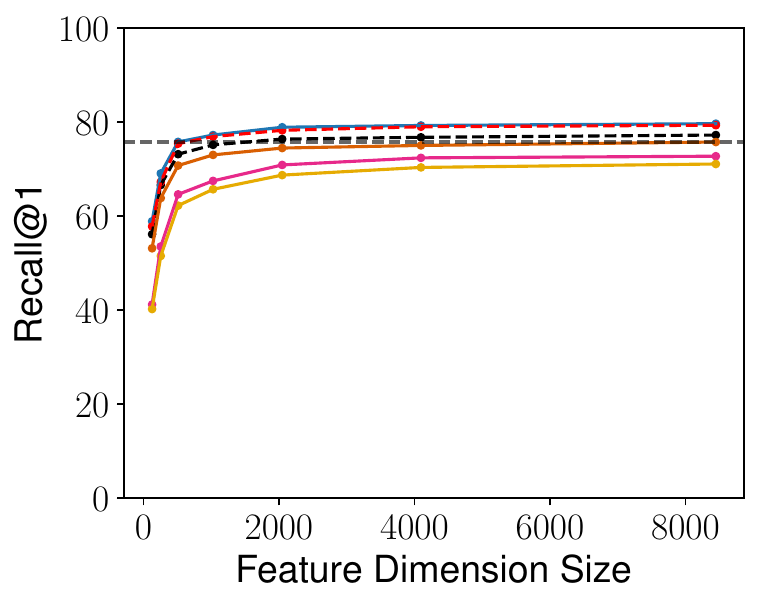} & 
        \includegraphics[width=0.4625\linewidth]{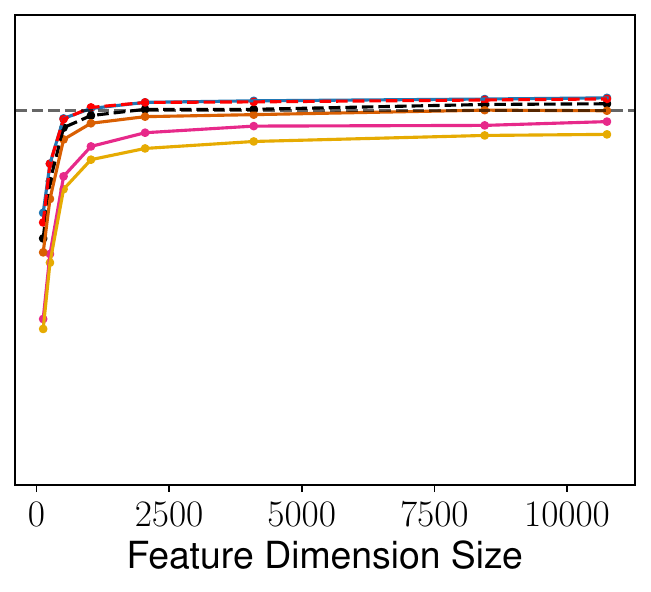} \\
        \small{\textbf{EigenPlaces (512D)}} & \small{\textbf{NetVLAD (4096D)}} \\
        \includegraphics[width=0.5375\linewidth]{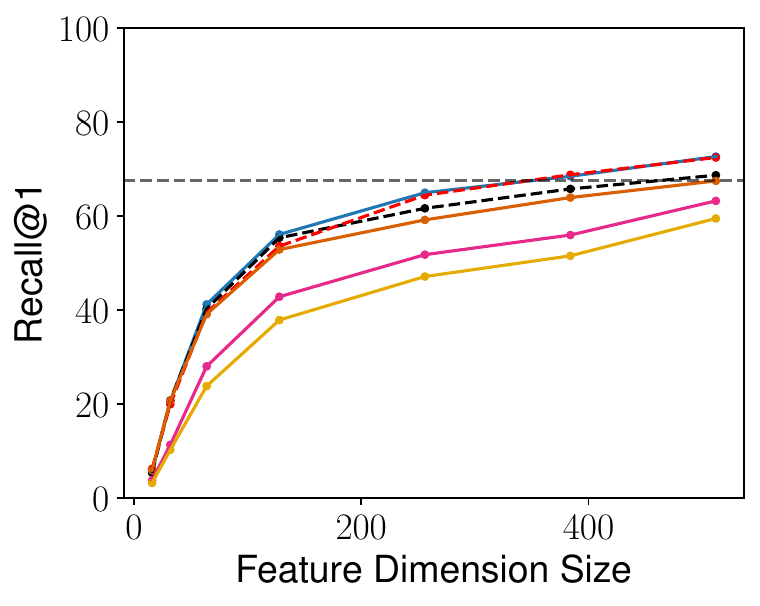} & 
        \includegraphics[width=0.4625\linewidth]{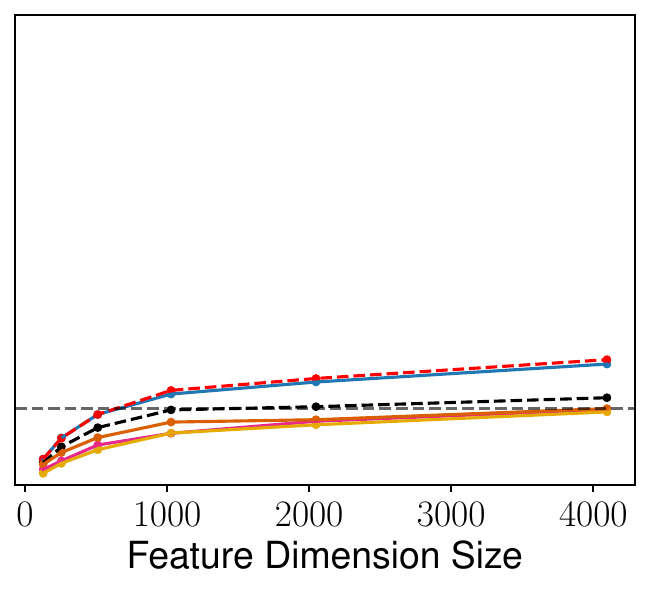} \\
        \multicolumn{2}{r}{\includegraphics[width=\linewidth]{figures/supp/ftr_reduction/Nord/ftr_red_abl_legend_v5.pdf}}
    \end{tabular}
    \caption{Recall@1 performance for different VPR descriptors across the Nordland Winter set as dimensionality is reduced using Gaussian Random Projection.}
    \label{fig:nord_winter_ftr_red}
\end{figure}

\begin{figure}[h!t]
    \centering
    \setlength\tabcolsep{1.2mm}
    \begin{tabular}{cc}
        \multicolumn{2}{c}{\textbf{Query: SFU Mountain Wet}} \\
        \small{\textbf{CosPlace (512D)}} & \small{\textbf{MixVPR (4096D)}} \\
        \includegraphics[width=0.5375\linewidth]{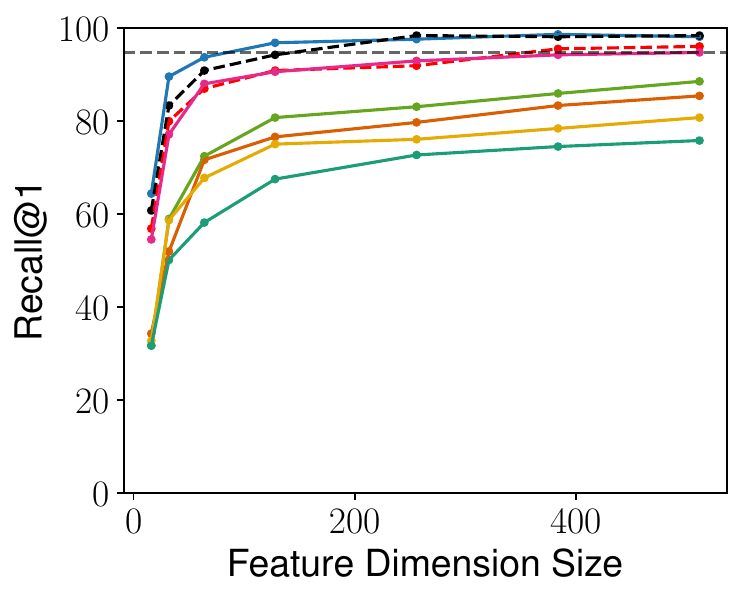} &  
        \includegraphics[width=0.4625\linewidth]{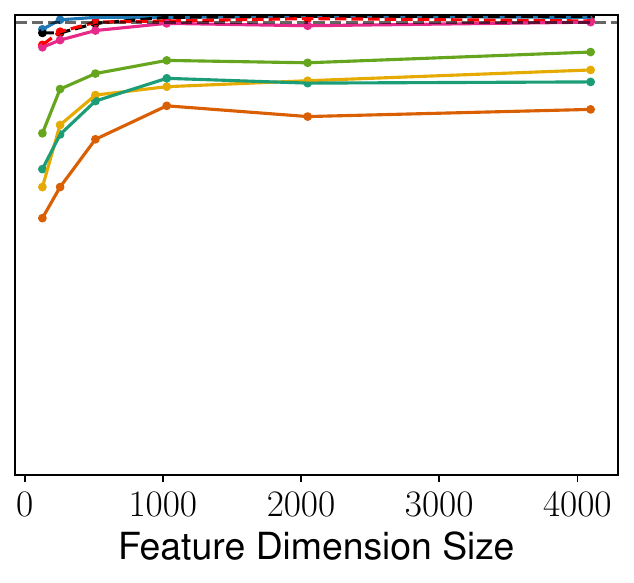} \\
        \small{\textbf{SALAD (8448D)}} & \small{\textbf{CricaVPR (10752D)}} \\
        \includegraphics[width=0.5375\linewidth]{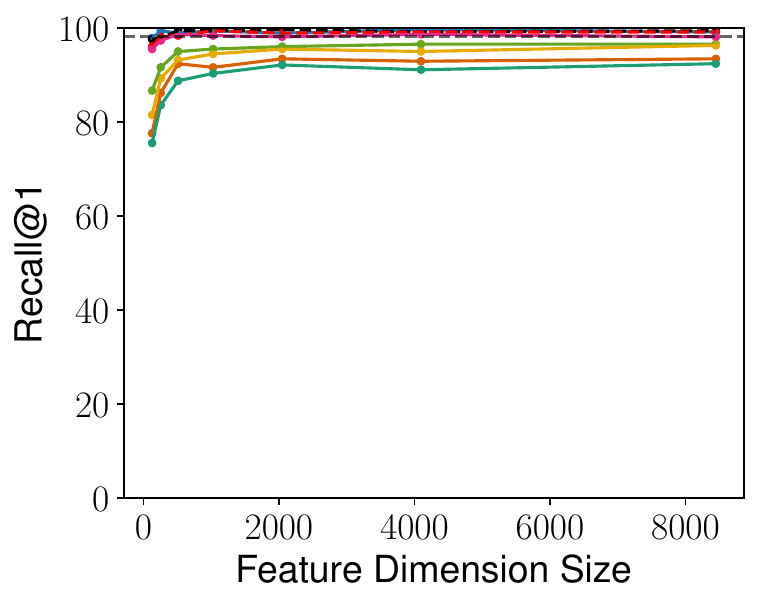} & 
        \includegraphics[width=0.4625\linewidth]{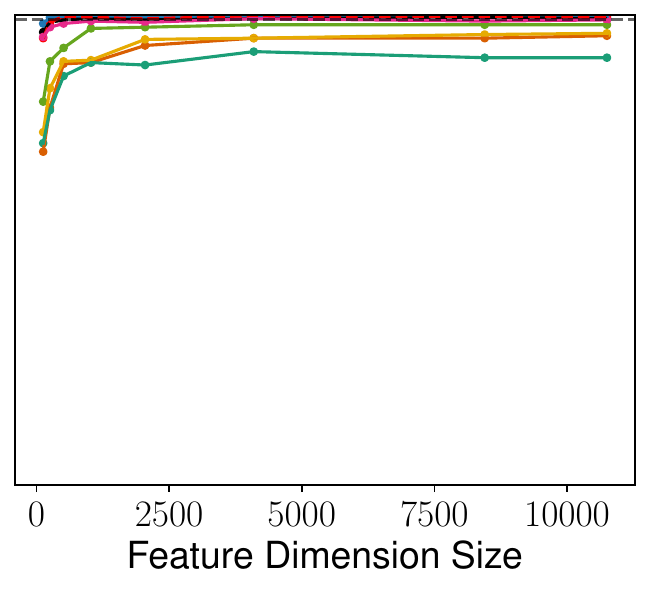} \\
        \small{\textbf{EigenPlaces (512D)}} & \small{\textbf{NetVLAD (4096D)}} \\
        \includegraphics[width=0.5375\linewidth]{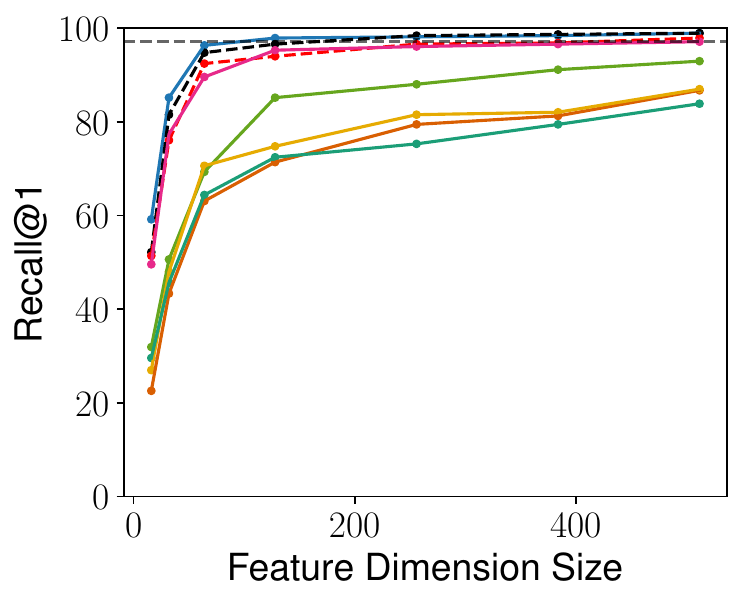} & 
        \includegraphics[width=0.4625\linewidth]{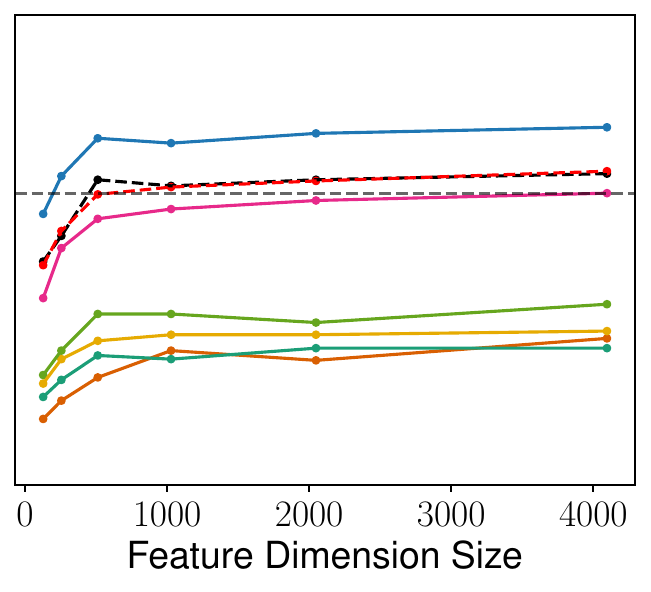} \\
        \multicolumn{2}{r}{\includegraphics[width=\linewidth]{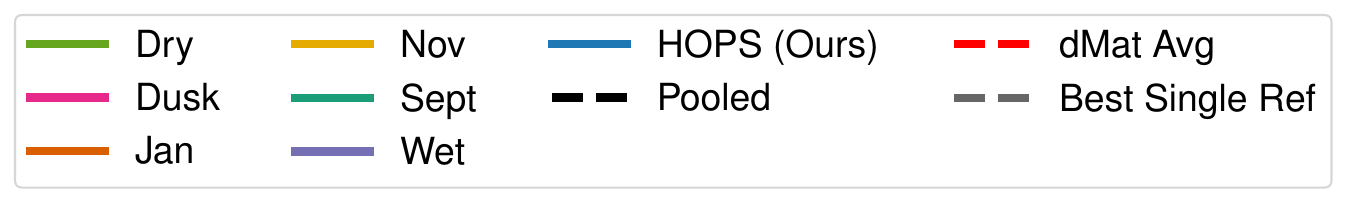}}
    \end{tabular}
    \caption{Recall@1 performance for different VPR descriptors across the SFU Mountain Wet set as dimensionality is reduced using Gaussian Random Projection.}
    \label{fig:sfu_wet_ftr_red}
\end{figure}

\begin{figure}[h!t]
    \centering
    \setlength\tabcolsep{1.2mm}
    \begin{tabular}{cc}
        \multicolumn{2}{c}{\textbf{Query: SFU-Mountain January}} \\
        \small{\textbf{CosPlace (512D)}} & \small{\textbf{MixVPR (4096D)}} \\
        \includegraphics[width=0.5375\linewidth]{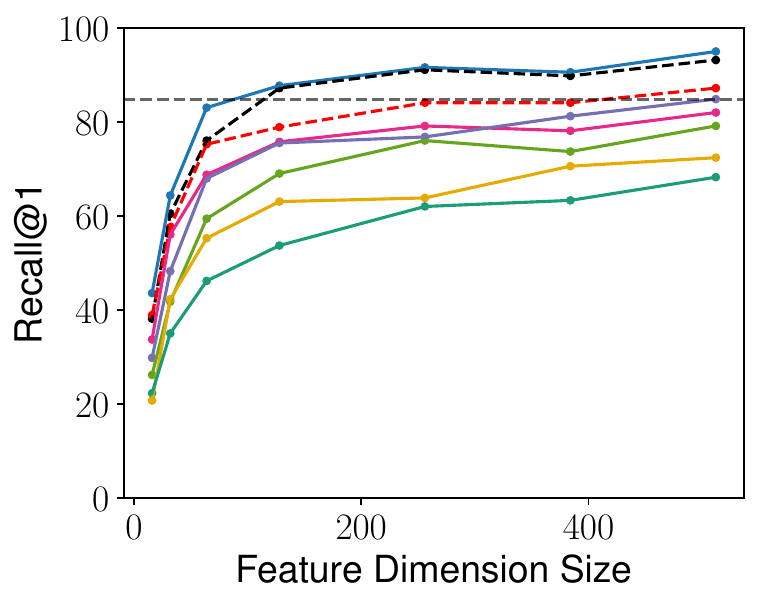} &  
        \includegraphics[width=0.4625\linewidth]{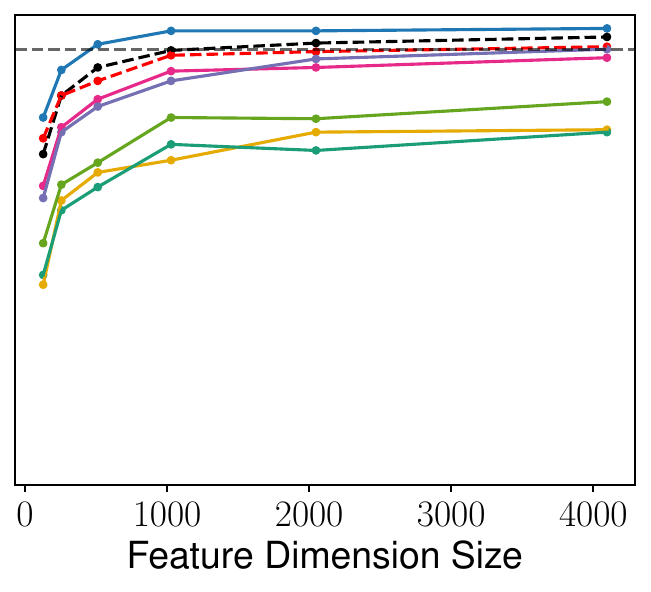} \\
        \small{\textbf{SALAD (8448D)}} & \small{\textbf{CricaVPR (10752D)}} \\
        \includegraphics[width=0.5375\linewidth]{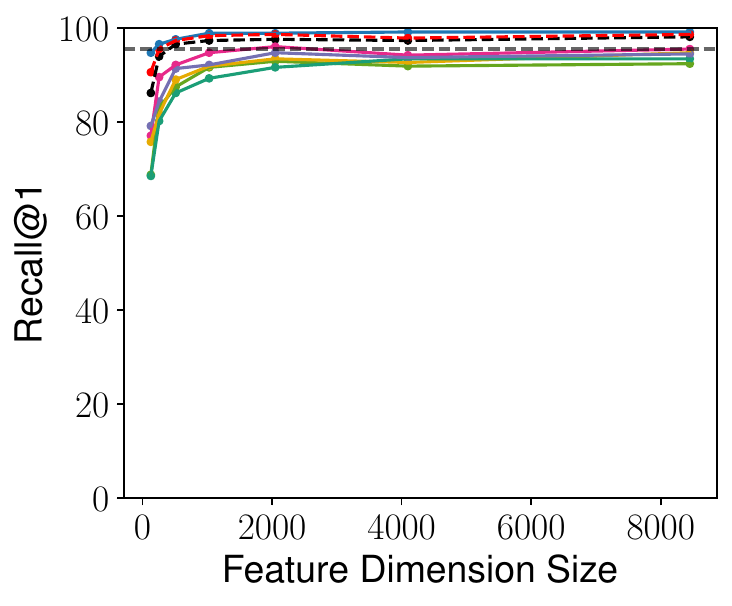} & 
        \includegraphics[width=0.4625\linewidth]{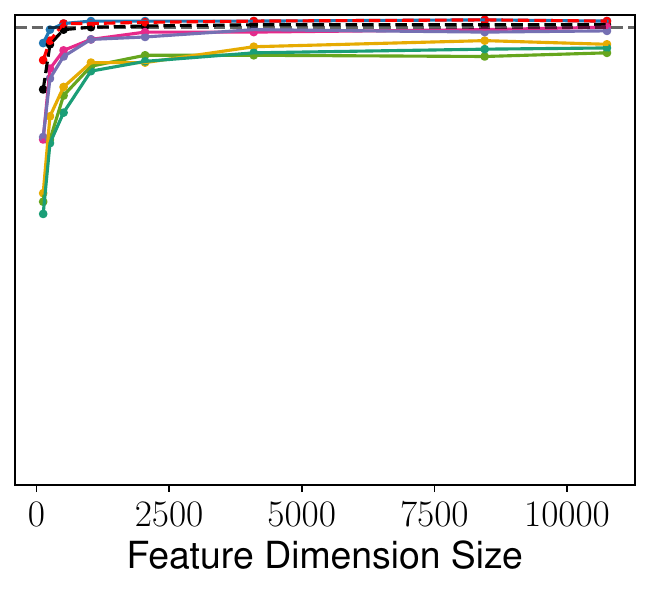} \\
        \small{\textbf{EigenPlaces (512D)}} & \small{\textbf{NetVLAD (4096D)}} \\
        \includegraphics[width=0.5375\linewidth]{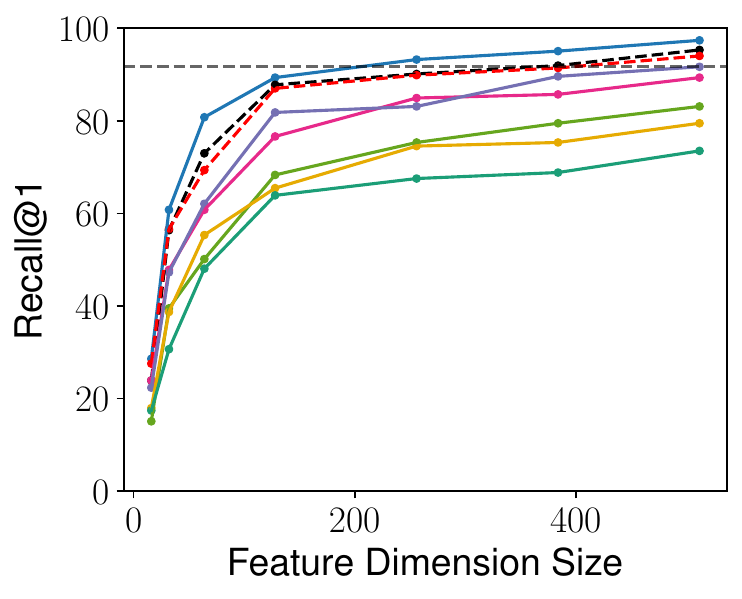} & 
        \includegraphics[width=0.4625\linewidth]{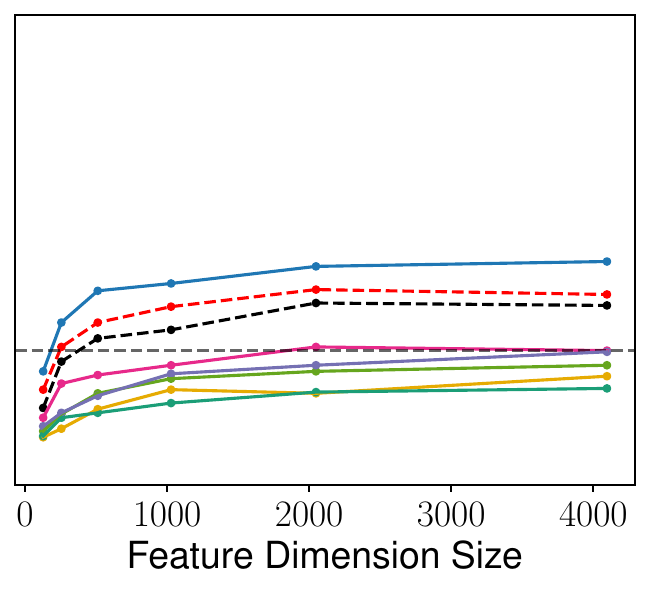} \\
        \multicolumn{2}{r}{\includegraphics[width=\linewidth]{figures/supp/ftr_reduction/SFU/ftr_red_abl_legend_v5.pdf}}
    \end{tabular}
    \caption{Recall@1 performance for different VPR descriptors across the SFU Mountain January set as dimensionality is reduced using Gaussian Random Projection.}
    \label{fig:sfu_jan_red}
\end{figure}

In the main manuscript, we investigated how recall@1 performance is affected by reducing the dimensionality of descriptors using a Gaussian Random Projection and presented results for the Oxford RobotCar Dusk query set. Here, we extend these results and provide the corresponding figures for all six VPR descriptors across two additional query sets from each dataset (Figures~\ref{fig:rc_night_ftr_red}--\ref{fig:sfu_jan_red}). The additional results confirm the finding of the main manuscript and demonstrate that our \textcolor{HOPS}{HOPS} fused descriptors can achieve the same or better performance compared to using the best single reference set, with up to a $97\%$ reduction in descriptor dimensionality. Despite the low dimensionality of the CosPlace~\cite{berton2022rethinking} and EigenPlaces~\cite{berton2023eigenplaces} descriptors (512D), our \textcolor{HOPS}{HOPS} fused descriptors still match or exceed the best single reference recall with reduced dimensionality for $9$/$12$ cases shown.

\clearpage

\section{Unstructured Datasets: Google Landmarks}
\label{sec:supp_unstruct_dsets}
One of the major challenges in robotics, and VPR, is maintaining performance in dynamic and unstructured environments. Datasets and results included in the main manuscript already contain extensive instances of dynamic environments, with environmental condition changes evident in all datasets. For example, the Oxford RobotCar~\cite{RobotCarDatasetIJRR} dataset contains weather/time of day changes, dynamic objects (i.e. cars and pedestrians), and temporal changes such as construction and infrastructure changes. All of these datasets contain images captured from the same route throughout an environment which could be considered independently as either query or reference sets. However, some datasets are unstructured and contain a collection of non-sequential images, with a varied number of images per `place', which can be localized using meta data such as latitude, longitude, and heading.

In this section, we present and discuss results for using our \textcolor{HOPS}{HOPS} fused descriptors to compress the more unstructured reference dataset for the \hyperlink{https://www.kaggle.com/datasets/confirm/google-landmark-dataset-v2-micro}{Google Landmarks v2 micro} dataset. This version of the Google Landmarks v2 dataset contains $23,294$ reference images and $3,103$ query images. Using our \textcolor{HOPS}{HOPS} approach, we were able to fuse 7-9 reference images from each place and reduce the reference set to just $3,103$ images; $13.3\%$ of the original size. Using the SALAD VPR descriptor~\cite{izquierdo2024optimal}, Table~\ref{table:supp_google_lnd} shows that \textcolor{HOPS}{HOPS} fused descriptors are able to substantially reduce the reference set size while only incurring a small decrease in Recall@1 performance ($3.9\%$), therefore \textit{significantly} reducing compute and storage requirements.

For comparison, we also provide results for an alternative strategy where dimensionality reduction is used to reduce the reference set size rather than our \textcolor{HOPS}{HOPS} descriptors. We reduce feature vectors to 1024 dimensional to provide equivalent memory requirements compared to our \textcolor{HOPS}{HOPS} reduced reference set. Table~\ref{table:supp_google_lnd} shows that our \textcolor{HOPS}{HOPS} fused descriptors are able to maintain a much higher Recall@1 at this memory footprint compared to using dimensionality reduction methods.

\begin{table}[h!]
    \caption{Reference set attributes and Recall@1 performance using SALAD~\cite{izquierdo2024optimal} on the Google Landmarks v2 micro dataset for different reference set reduction strategies. Our \textcolor{HOPS}{HOPS} fused descriptors significantly reduce reference set size while only incurring a small decrease in Recall@1.}
    \centering
    \footnotesize
    \vspace{-0.2cm}
    \begin{tabular}{l*{2}{c}|c}
        \toprule
        \textbf{Reference Set} 
        & \textbf{Original} 
        & \textbf{\textcolor{HOPS}{HOPS} Reduced}
        & \textbf{Dim-Reduced Feats.\ }\\
        \hline
        Num. of Refs. & 23,294 & 3,103 & 23,294 \\
        Feat. Dim. & 8488 & 8448 & 1024 \\
        Recall@1      & 69.7   & 65.8 & 59.7 \\
        \bottomrule
    \end{tabular}
    \label{table:supp_google_lnd}
\end{table}

\section{Dataset Identification}
\label{sec:supp_dset_id}

\begin{figure*}[!ht]
  \centering

  \includegraphics[width=\linewidth,trim={1mm 1mm 1mm 1mm},clip]{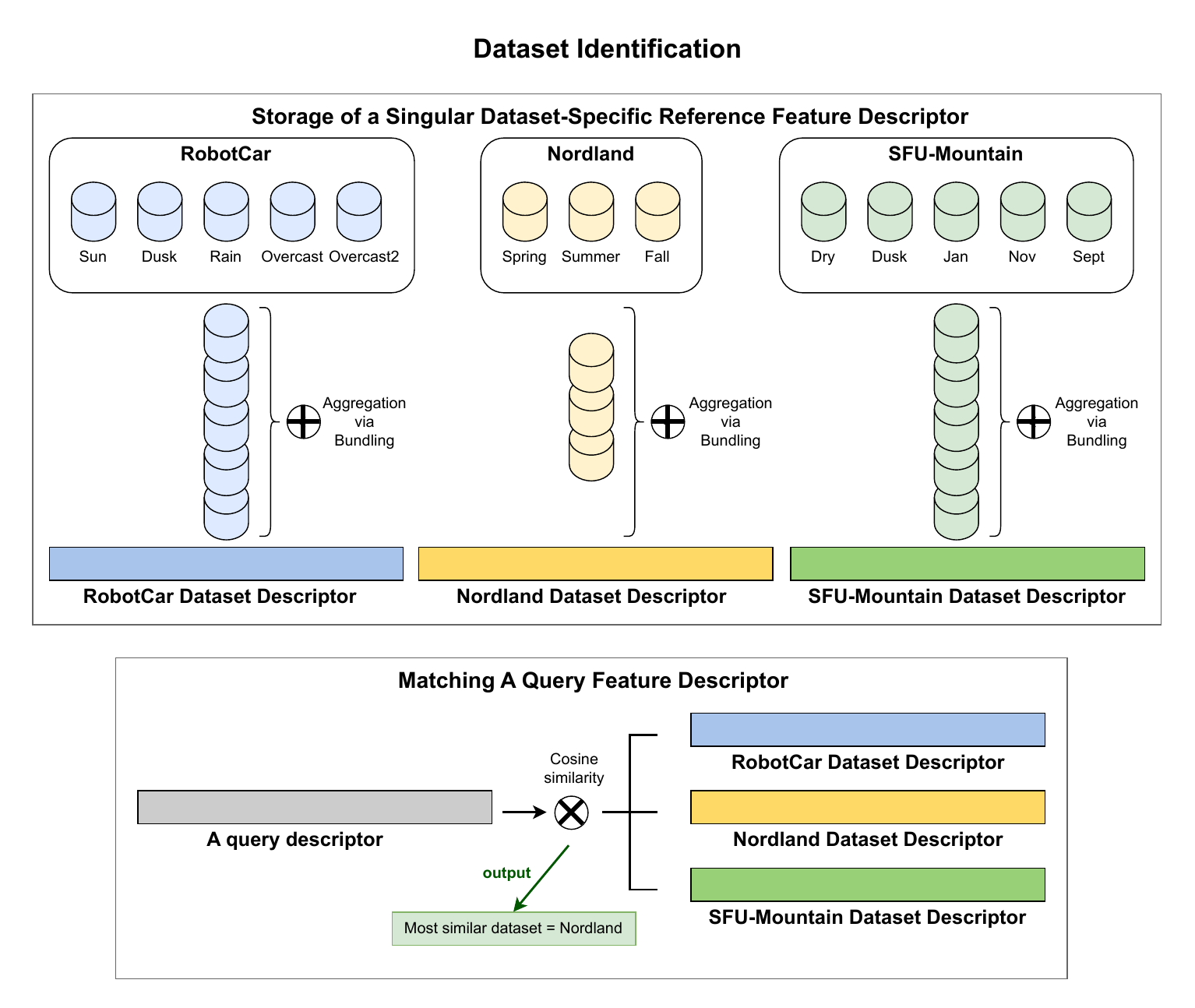}

  \vspace*{-0.4cm}
  \caption{
  The visualisation of our dataset identification investigation, as discussed earlier in Section~\ref{subsec:other_apps}. 
  \textbf{Top:} We used our \textcolor{HOPS}{HOPS} fused descriptors to aggregate the reference descriptors from each dataset into a single and unified feature descriptor to represent the entire dataset, essentially summarizing each dataset via a single distinct representation. 
  \textbf{Bottom:} To classify a query descriptor, we computed its cosine similarity to each reference dataset descriptor. The reference dataset descriptor with the highest cosine similarity is identified as the predicted dataset to which the query belongs. 
  }
  \label{fig:dataset_identification}
  \vspace*{-0.2cm}
\end{figure*}

Beyond fusing descriptors from the same place, there are many other possible applications for the HDC framework in VPR, such as dataset/environment identification. This section provides additional details on using our \textcolor{HOPS}{HOPS} fused descriptors to identify which dataset a given query descriptor belongs to, as discussed in Section~\ref{subsec:other_apps} of the main manuscript. Figure~\ref{fig:dataset_identification} provides a visualization of the overall process we used to perform this experiment. 

First, for each dataset, we pool all respective available reference sets together. Then, we use the HDC bundling operation to aggregate \textit{all} descriptors into a \textit{single overall dataset descriptor}. After separately performing the bundling for the three datasets, RobotCar, Nordland, and SFU Mountain, this provides three `dataset' reference descriptors.

To evaluate the performance of our \textit{dataset-specific fused descriptor}, we identified the source dataset of each query descriptor by calculating its cosine similarity against each of the single overall dataset descriptors. We emphasize that when evaluating the accuracy of dataset identification for each query set, the respective set was removed from the bundling operation and therefore was not included in the overall dataset descriptors.

To provide context for these results, we compare against alternate dataset identification approaches. %
The first method we compare to is one where a \textit{single} descriptor is randomly chosen to represent each dataset from the respective reference sets. We also evaluate dataset identification accuracy when $10$ descriptors are randomly chosen per dataset, covering multiple reference sets. For this approach, the query descriptor is compared to all $10$ reference descriptors from each dataset and the predicted dataset becomes the one which contains the reference descriptor most similar to the query according to cosine similarity.

Table~\ref{table:dset_id} provides the accuracy of our \textcolor{HOPS}{HOPS} fused descriptors and all comparisons for the dataset identification task using the SALAD VPR descriptor. It demonstrates that our approach can correctly predict the source dataset of a query image, with an accuracy of above $99.7\%$ across all datasets, which is significantly better than a random single image ($20.5\%$ on average) and 10 random images per dataset ($2.6\%$ on average).%

This experiment shows that our \textcolor{HOPS}{HOPS} fused descriptor is able to distinguish between different environments based on their overall feature descriptor characteristics.
\clearpage
\flushbottom
\setlength\tabcolsep{1.0mm} %
\begin{table*}
    \caption{Dataset identification accuracy across all datasets using the SALAD VPR descriptor.}
    \centering
    \small
    \begin{tabular}{l*{5}{c}|*{4}{c}|*{6}{c}}
        \toprule
         &
        \multicolumn{5}{c}{\textbf{Oxford RobotCar}} & \multicolumn{4}{c}{\textbf{Nordland}} & \multicolumn{6}{c}{\textbf{SFU-Mountain}}\\
        \hline
        \textbf{Queries $\rightarrow$} &
        \rotatebox{90}{\raisebox{0.075cm}{\textbf{Dusk}}} & \rotatebox{90}{\raisebox{0.075cm}{\textbf{Night}}} & \rotatebox{90}{\raisebox{0.075cm}{\textbf{Overcast}}} & \rotatebox{90}{\raisebox{0.075cm}{\textbf{Overcast2\ }}} & \rotatebox{90}{\raisebox{0.075cm}{\textbf{Rain}}} &
        \rotatebox{90}{\raisebox{0.075cm}{\textbf{Fall}}} & \rotatebox{90}{\raisebox{0.075cm}{\textbf{Spring}}} & \rotatebox{90}{\raisebox{0.075cm}{\textbf{Summer}}} & \rotatebox{90}{\raisebox{0.075cm}{\textbf{Winter}}} &
        \rotatebox{90}{\raisebox{0.075cm}{\textbf{Dry}}} & \rotatebox{90}{\raisebox{0.075cm}{\textbf{Dusk}}} & \rotatebox{90}{\raisebox{0.075cm}{\textbf{Jan}}} & \rotatebox{90}{\raisebox{0.075cm}{\textbf{Nov}}} & \rotatebox{90}{\raisebox{0.075cm}{\textbf{Sept}}} & \rotatebox{90}{\raisebox{0.075cm}{\textbf{Wet}}} \\
        \hline
        \textbf{Single Image} & 95.54 &  62.41 &  63.67 &  84.49 &  76.37 &  
        79.40 &  94.89 &  8.91 &  64.30 &  
        85.97 &  97.92 &  98.96 &   100 &  96.10 &  97.14 \\
        \textbf{Pooled (10 Images)} & 94.61 &  96.88 &  93.19 &  95.82 &  97.65 &  
        95.19 &  96.96 &  94.72 &  94.84 &  
         100 &   100 &   100 &   100 &   100 &   100 \\
        \textbf{\textcolor{HOPS}{HOPS} (Ours)} & 100 & 100 & 99.85 & 99.79 & 99.79 &
        99.97 & 99.82 & 99.87 & 99.8 & 
        100 & 100 & 99.74 & 100 & 100 & 100 \\
        \bottomrule
    \end{tabular}
    \label{table:dset_id}
\end{table*}

\clearpage

\section{Other Feature Aggregation Methods}
\label{sec:supp_feat_agg}
In the main manuscript, we compare our \textcolor{HOPS}{HOPS} fused descriptors to the distance matrix averaging feature aggregation approach because it was the highest performing method from~\cite{fischer2020event}. In this section, we provide a comparison to the other feature aggregation methods explored in~\cite{fischer2020event}. These include taking the \textit{minimum} values, \textit{maximum} values, or \textit{median} values from the distance matrix rather than the mean/average. We present results obtained using MixVPR~\cite{ali2023mixvpr} VPR descriptors on the Oxford RobotCar datasets~\cite{RobotCarDatasetIJRR}.

Similarly to results seen in~\cite{fischer2020event}, Table~\ref{table:supp_feat_agg} shows that the distance matrix averaging generally achieves equivalent or higher Recall@1 compared to the other feature aggregation methods from~\cite{fischer2020event}. Notably, the `Minimum' method achieves similar results on the `Overcast' and `Rain' datasets but considerably lower Recall@1 for more challenging conditions such as `Dusk' and `Rain'. Importantly, our \textcolor{HOPS}{HOPS} fused descriptors generally maintain the highest Recall@1 out of all methods.

We would also like to reiterate the key advantages of \textcolor{HOPS}{HOPS} compared to these other feature aggregation methods. \textcolor{HOPS}{HOPS} fuses place descriptors while all methods from ~\cite{fischer2020event} fuse difference matrices obtained by running VPR on $N$ reference-query image pairs per place; requiring to store $N$ descriptors per place. \cite{fischer2020event}’s vanilla version and \cite{neubert2021hyperdimensional} need to compute $N$ query image descriptors at inference, giving \textcolor{HOPS}{HOPS} both significant computation (single query descriptor) and memory (single reference vector) advantages, and higher performance.

\begin{table}[h!]
    \caption{Recall@1 on the RobotCar datasets using MixVPR and different feature aggregation methods. In addition to methods compared in the main manuscript, we present other approaches also explored in~\cite{fischer2020event}. The results show the dMat averaging aggregation generally achieves higher Recall@1 across the different datasets compared to other methods from~\cite{fischer2020event}; with \textcolor{HOPS}{HOPS} nearly always achieving the highest Recall@1 of all methods.}
    \centering
    \scriptsize
    \vspace{-0.2cm}
    \begin{tabular}{l*{9}{c}|*{3}{c}}
        \toprule
        \textbf{References} 
        & \rotatebox{90}{\raisebox{0.075cm}{\textbf{Sun}}} 
        & \rotatebox{90}{\raisebox{0.075cm}{\textbf{Dusk}}} 
        & \rotatebox{90}{\raisebox{0.075cm}{\textbf{Night}}} 
        & \rotatebox{90}{\raisebox{0.075cm}{\textbf{O/C}}} 
        & \rotatebox{90}{\raisebox{0.075cm}{\textbf{O/C2}}} 
        & \rotatebox{90}{\raisebox{0.075cm}{\textbf{Rain}}} 
        & \rotatebox{90}{\raisebox{0.075cm}{\textbf{Avg~[22]}}} 
        & \rotatebox{90}{\raisebox{0.075cm}{\textbf{Pool}}} 
        & \rotatebox{90}{\raisebox{0.075cm}{\textbf{\textcolor{HOPS}{HOPS}}}} 
        & \rotatebox{90}{\raisebox{0.075cm}{\textbf{Min~[22]}}}
        & \rotatebox{90}{\raisebox{0.075cm}{\textbf{Max~[22]}}}
        & \rotatebox{90}{\raisebox{0.075cm}{\textbf{Med~[22]}}}\\
        \hline
        \textbf{Queries $\downarrow$} & \multicolumn{9}{c}{\textbf{MixVPR (4096D)}}\\
        \hline
        Dusk      & 69.0   & -    & 64.6 & \underline{71.7} & 67.4 & 68.3 & 82.9 & 77.1 & \textbf{83.1} & 77.1 & 65.2 & 76.7 \\
        Night     & 50.9 & \underline{59.2} & -    & 57.2 & 52.0   & 46.0   & \textbf{70.0}   & 60.1 & 68.8 & 60.1 & 53.8 & 60.3 \\
        Overcast  & 86.3 & 60.1 & 52.2 & -    & \underline{89.1} & 87.1 & 91.5 & 92.0   & \textbf{93.3} & 92.0 & 52.6 & 85.9 \\
        Overcast2 & \underline{91.2} & 61.4 & 50.3 & 90.6 & -    & 88.8 & 93.6 & 93.7 & \textbf{94.7} & 93.7 & 51.6 & 88.8 \\
        Rain      & 88.7 & 63.6 & 48.3 & \underline{89.6} & 89.5 & -    & 92.7 & 93.4 & \textbf{94.7} & 93.4 & 50.8 & 88.1 \\
        \bottomrule
    \end{tabular}
    \label{table:supp_feat_agg}
\end{table}

\section{Qualitative Results}
\label{sec:qual_results}

\begin{figure*}[t]
  \centering

  \subfloat{\includegraphics[width=\linewidth,trim={1mm 1mm 1mm 1mm},clip]{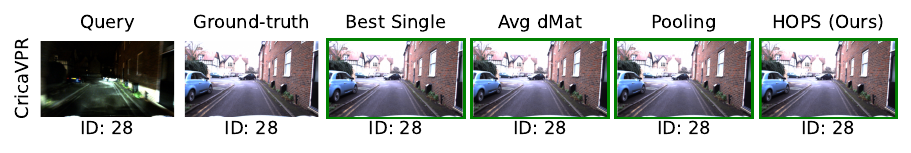}}

  \subfloat{\includegraphics[width=\linewidth,trim={1mm 1mm 1mm 6mm},clip]{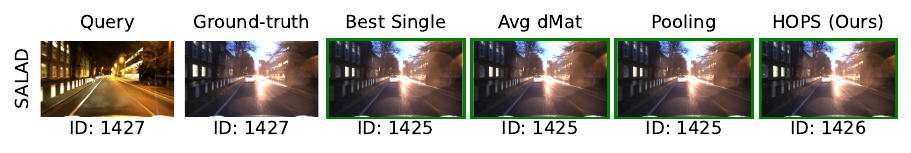}}

  \subfloat{\includegraphics[width=\linewidth,trim={1mm 1mm 1mm 6mm},clip]{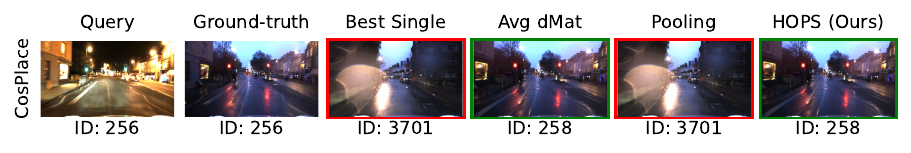}}
  
  \subfloat{\includegraphics[width=\linewidth,trim={1mm 1mm 1mm 6mm},clip]{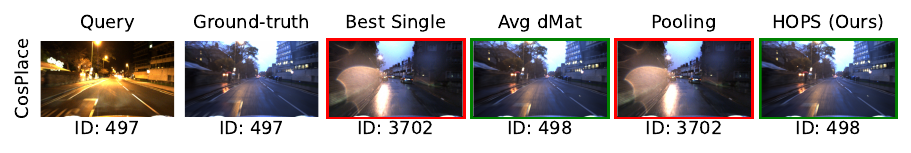}}

  \subfloat{\includegraphics[width=\linewidth,trim={1mm 1mm 1mm 6mm},clip]{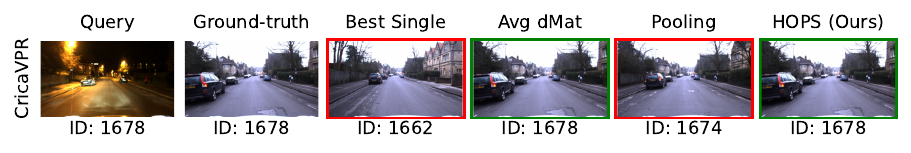}}
  
  \subfloat{\includegraphics[width=\linewidth,trim={1mm 1mm 1mm 6mm},clip]{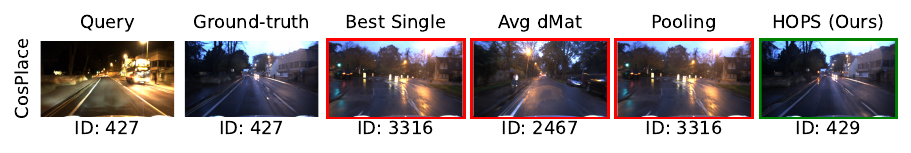}}

  \subfloat{\includegraphics[width=\linewidth,trim={1mm 1mm 1mm 6mm},clip]{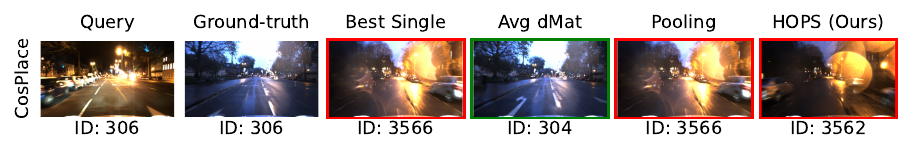}}

  \subfloat{\includegraphics[width=\linewidth,trim={1mm 1mm 1mm 6mm},clip]{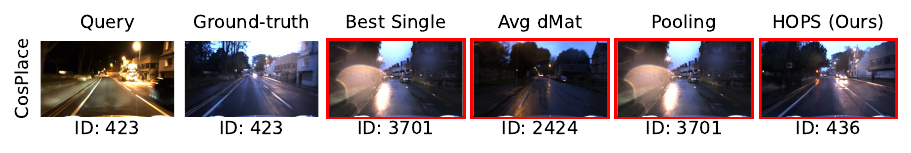}}

  \vspace*{-0.4cm}
  \caption{Qualitative results of our method (\textcolor{HOPS}{HOPS}) on the RobotCar dataset with Night set as the query. 
  For each row, the displayed reference set corresponds to the `Best Single' reference set for the specific method, shown to the right of the query image. 
  We show three different scenarios:
  i. Cases where fusing reference sets via \textcolor{HOPS}{HOPS} (ours) produces correct matches, similar to other techniques (rows 1 and 2). 
  ii. Cases where \textcolor{HOPS}{HOPS} produces correct matches while at least one other method fails (rows 3, 4, 5, and 6).
  iii. Cases where \textcolor{HOPS}{HOPS} retrieves false matches and other methods either succeed or fail (rows 7 and 8). 
  Note that we use a tight ground truth tolerance of $\pm$ 2 meters for the RobotCar dataset. \textcolor{HOPS}{HOPS} fused descriptors further reduce the error of matches already made in close proximity to the true match, effectively disambiguating spatially close places. 
  }
  \label{fig:RobotCar_qualitative_results}
  \vspace*{-0.2cm}
\end{figure*}

\begin{figure*}[t]
  \centering

  \subfloat{\includegraphics[width=\linewidth,trim={1mm 1mm 1mm 1mm},clip]{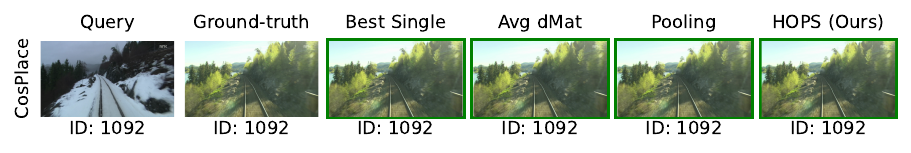}}

  \subfloat{\includegraphics[width=\linewidth,trim={1mm 1mm 1mm 6mm},clip]{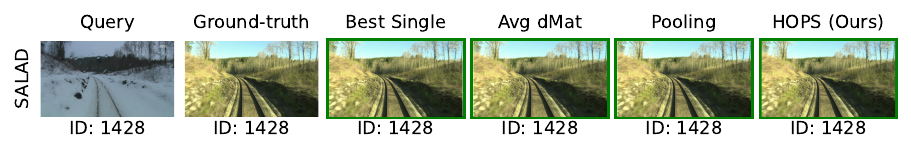}}

  \subfloat{\includegraphics[width=\linewidth,trim={1mm 1mm 1mm 6mm},clip]{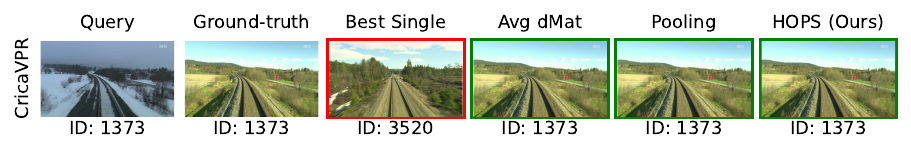}}

  \subfloat{\includegraphics[width=\linewidth,trim={1mm 1mm 1mm 6mm},clip]{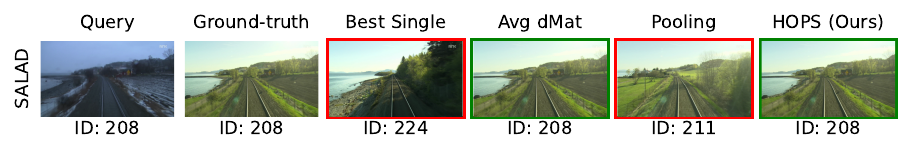}}
  
  \subfloat{\includegraphics[width=\linewidth,trim={1mm 1mm 1mm 6mm},clip]{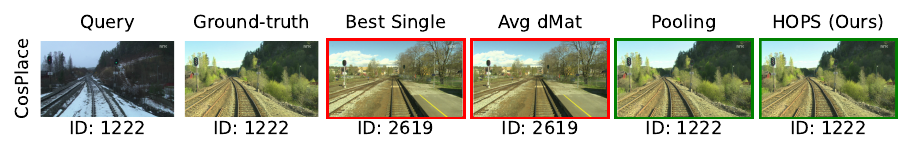}}

  \subfloat{\includegraphics[width=\linewidth,trim={1mm 1mm 1mm 6mm},clip]{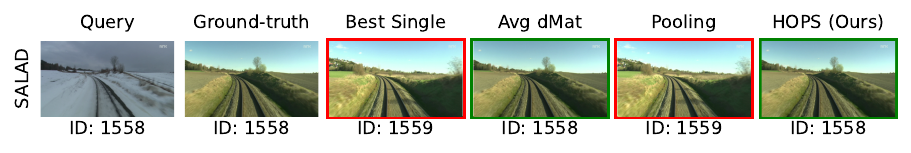}}
  
  \subfloat{\includegraphics[width=\linewidth,trim={1mm 1mm 1mm 6mm},clip]{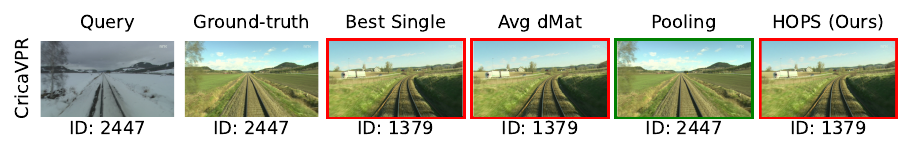}}

  \subfloat{\includegraphics[width=\linewidth,trim={1mm 1mm 1mm 6mm},clip]{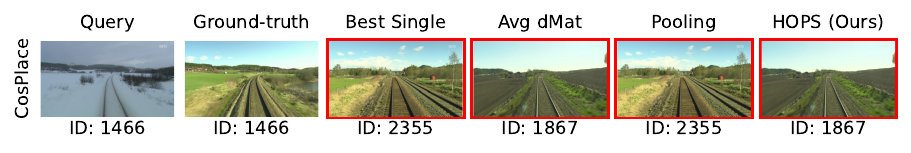}}

  \vspace*{-0.4cm}
  \caption{Qualitative results of our method (\textcolor{HOPS}{HOPS}) on the Nordland dataset with Winter set as the query. 
  For each row, the displayed reference set corresponds to the `Best Single' reference set for the specific method, shown to the right of the query image. 
  We show three different scenarios:
  i. Cases where fusing reference sets via \textcolor{HOPS}{HOPS} (ours) produces correct matches, similar to other techniques (rows 1 and 2). 
  ii. Cases where \textcolor{HOPS}{HOPS} produces correct matches while at least one other method fails (rows 3, 4, 5, and 6).
  iii. Cases where \textcolor{HOPS}{HOPS} retrieves false matches and other methods either succeed or fail (rows 7 and 8). 
  Note that we use a tight ground truth tolerance of $0$ images for the Nordland dataset. }
  \label{fig:Nordland_qualitative_results}
  \vspace*{-0.2cm}
\end{figure*}

\begin{figure*}[t]
  \centering

  \subfloat{\includegraphics[width=\linewidth,trim={1mm 1mm 1mm 1mm},clip]{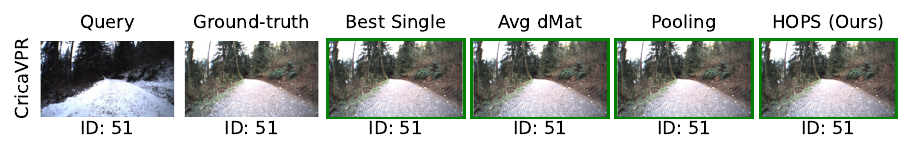}}

  \subfloat{\includegraphics[width=\linewidth,trim={1mm 1mm 1mm 6mm},clip]{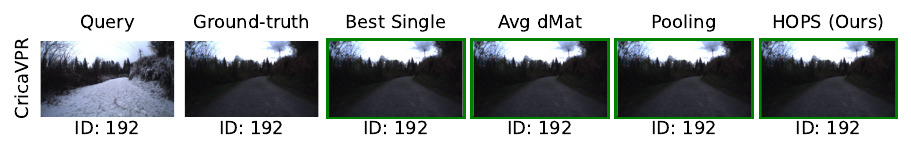}}

  \subfloat{\includegraphics[width=\linewidth,trim={1mm 1mm 1mm 6mm},clip]{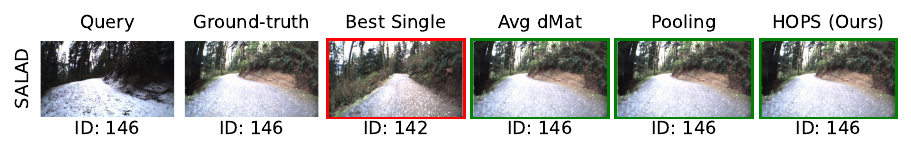}}

  \subfloat{\includegraphics[width=\linewidth,trim={1mm 1mm 1mm 6mm},clip]{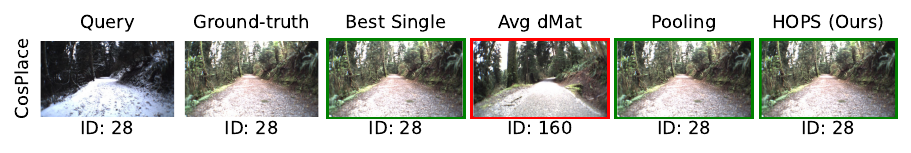}}

  \subfloat{\includegraphics[width=\linewidth,trim={1mm 1mm 1mm 6mm},clip]{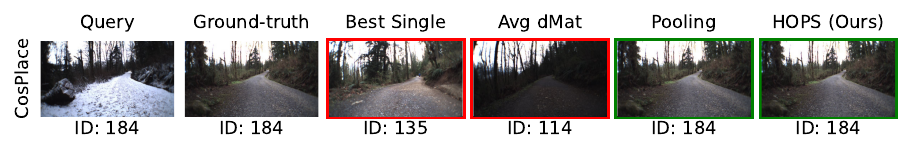}}
  
  \subfloat{\includegraphics[width=\linewidth,trim={1mm 1mm 1mm 6mm},clip]{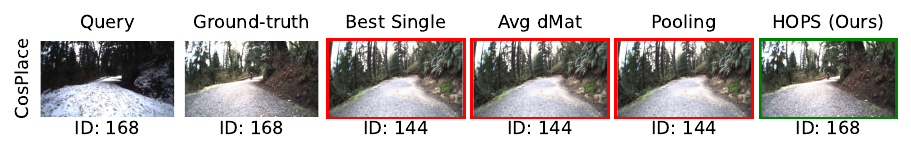}}

  \subfloat{\includegraphics[width=\linewidth,trim={1mm 1mm 1mm 6mm},clip]{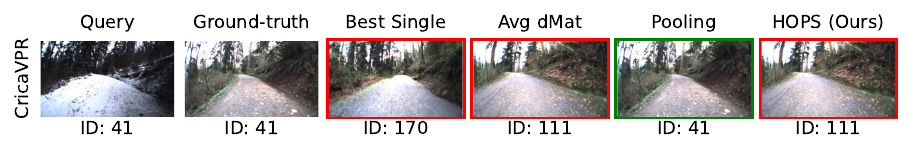}}
  
  \subfloat{\includegraphics[width=\linewidth,trim={1mm 1mm 1mm 6mm},clip]{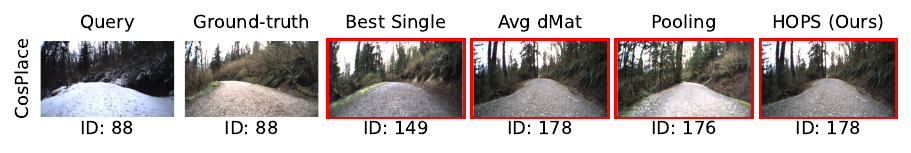}}

  \vspace*{-0.4cm}
  \caption{Qualitative results of our method (\textcolor{HOPS}{HOPS}) on the SFU-Mountain dataset with January set as the query. 
  For each row, the displayed reference set corresponds to the `Best Single' reference set for the specific method, shown to the right of the query image. 
  We show three different scenarios:
  i. Cases where fusing reference sets via \textcolor{HOPS}{HOPS} (ours) produces correct matches, similar to other techniques (rows 1 and 2). 
  ii. Cases where \textcolor{HOPS}{HOPS} produces correct matches while at least one other method fails (rows 3, 4, 5, and 6).
  iii. Cases where \textcolor{HOPS}{HOPS} retrieves false matches and other methods either succeed or fail (rows 7 and 8). 
  Note that we use a tight ground truth tolerance of $\pm$ 1 image for the SFU-Mountain dataset.}
  \label{fig:SFU_Mountain_qualitative_results}
  \vspace*{-0.2cm}
\end{figure*}

In this section, we provide qualitative results to show scenarios where our method excels and instances where it fails, evaluated across a range of diverse and challenging conditions.
Figures~\ref{fig:RobotCar_qualitative_results},~\ref{fig:Nordland_qualitative_results}, and~\ref{fig:SFU_Mountain_qualitative_results} show qualitative results on the RobotCar, Nordland and SFU Mountain datasets, respectively. We use green borders to indicate correct matches, and red borders to indicate false matches. In these figures, we show cases where our \textcolor{HOPS}{HOPS} fused descriptors are able to retrieve correct matches even in cases where all other methods fail. We also include cases where \textcolor{HOPS}{HOPS} fails to retrieve a correct match, while other methods either retrieve correct matches or also fail. 

For comparability, we visualize VPR matches for all approaches using corresponding images from the single reference set which achieves the best recall@1; noting that for multi-reference methods such as the distance matrix averaging and our \textcolor{HOPS}{HOPS} fused descriptors, the matches rely on multiple reference sets.

We reiterate that we use a tight ground truth tolerance of $\pm$ 2 frames for the RobotCar dataset, $0$ frames for the Nordland dataset, and $\pm$ 1 frames for the SFU Mountain dataset. 
Therefore, while VPR methods are generally good at finding matches close to the ground truth location, our \textcolor{HOPS}{HOPS} fused descriptors are able to further reduce the match errors, even those near the true match, disambiguating spatially close places. 
For example, this improvement is evident in the RobotCar dataset in Figure~\ref{fig:RobotCar_qualitative_results} rows 2 and 5, the Nordland dataset in Figure~\ref{fig:Nordland_qualitative_results} rows 4 and 6, and the SFU Mountain dataset in Figure~\ref{fig:SFU_Mountain_qualitative_results} rows 3 and 6.

We have also included example visualizations where all or the majority of methods fail to match to the correct places, which are mostly due to high visual similarity between geographically distant places. These instances are shown for RobotCar, Nordland and SFU Mountain datasets in rows 7 and 8 of Figures~\ref{fig:RobotCar_qualitative_results},~\ref{fig:Nordland_qualitative_results}, and~\ref{fig:SFU_Mountain_qualitative_results}, respectively. 
\clearpage

\end{document}